\documentclass[twoside,11pt]{article}
\usepackage{jair, theapa, rawfonts}
\usepackage{ amssymb }
\usepackage{ amsmath }
\usepackage{pifont}
\usepackage{graphicx,xcolor}
\usepackage{caption}
\usepackage{float}
\usepackage{subcaption}
\usepackage{multirow}
\usepackage{rotating}
\usepackage{subfig}

\jairheading{1}{2016}{1-30}{6/16}{9/16}
\ShortHeadings{Time-Sensitive Bayesian Information Aggregation for Crowdsourcing Systems}
{Venanzi, Guiver, Kohli \& Jennings}
\firstpageno{1}

\begin{document}

\author{\name Matteo Venanzi \email mavena@microsoft.com \\
       \addr Microsoft, London, 2 Whaterhouse Square,
         EC1N 2ST UK
       \AND
       \name John Guiver \email joguiver@microsoft.com \\
       \name Pushmeet Kohli \email pkohli@microsoft.com \\
       \addr Microsoft Research, Cambridge, 21 Station Road CB1 2FB UK
              \AND
       \name Nicholas R. Jennings \email n.jennings@imperial.ac.uk \\
       \addr Imperial College, London, South Kensington SW7 2AZ UK}

\newcommand{\xmark}{\footnotesize-}%

\title{Time-Sensitive Bayesian  Information Aggregation for Crowdsourcing Systems}

\maketitle

\begin{abstract}
\noindent  Crowdsourcing systems commonly face the problem of aggregating multiple judgments provided by potentially unreliable workers. In addition, several aspects of the design of efficient crowdsourcing processes, such as defining worker's bonuses, fair prices and time limits of the tasks, involve knowledge of the likely duration of the task at hand.
Bringing this together, in this work we introduce a new time--sensitive Bayesian aggregation method that simultaneously estimates a task's duration  and obtains reliable aggregations of crowdsourced judgments.  Our method, called BCCTime, builds on the key insight that the time taken by a worker to perform a  task is an important indicator of the likely quality of the produced judgment.
To capture this, BCCTime uses latent variables to represent the uncertainty about the workers' completion time, the tasks' duration and the workers' accuracy. %
To relate the quality of a judgment to the time a worker spends on a task, our model assumes that each task is completed within a latent time window within which all workers with a propensity to genuinely attempt the  labelling task (i.e., no spammers) are expected to submit their judgments. In contrast, workers with a lower propensity to valid labelling, such as spammers, bots or lazy labellers, are assumed to perform tasks considerably faster or slower than the time required by normal workers.
Specifically, we use efficient message-passing Bayesian inference to learn approximate posterior probabilities of (i) the confusion matrix of each worker, (ii) the propensity to valid labelling of each worker, (iii)  the unbiased duration of each task and  (iv) the true label of each task. Using two real-world public datasets for entity linking tasks, we show that BCCTime produces up to 11\% more accurate classifications and  up to $100\%$ more informative estimates of a task's duration compared to state--of--the--art methods.
\end{abstract}

\section{Introduction}

Crowdsourcing has emerged as an effective way to  acquire large amounts of data that enables the development of a variety of applications driven by machine learning, human computation and participatory sensing systems \cite{kamar2012combining,bernstein2010soylent,zilli2014hidden}.
Services such as Amazon Mechanical Turk\footnote{\texttt{www.mturk.com}} (AMT), oDesk\footnote{\texttt{www.odesk.com}} and CrowdFlower\footnote{\texttt{www.crowdflower.com}} have enabled  a number of applications to hire pools of human workers to  provide data to serve for training image annotation \cite{whitehill2009whose,welinder2010multidimensional}, galaxy classification\footnote{\texttt{www.galaxyzoo.org}} \cite{kamar2012combining} and information retrieval systems \cite{alonso2008crowdsourcing}.
In such applications, a central problem is to deal with the diversity of accuracy and speed that workers exhibit when performing crowdsourcing tasks.
As a result, due to the uncertainty over the reliability of individual crowd responses, many systems collect many judgments from different workers to achieve high confidence in the quality of their labels. However, this can incur a high cost either in time or money, particularly when the workers are paid per judgment, or when a delay in the completion of the entire crowdsourcing project is introduced when workers intentionally delay their submissions to follow their own work schedule.
For example, in a typical crowdsourcing scenario, a requester must specify the number of requested assignments (i.e., individual responses from different workers), as well as the time limit for the completion of each assignment.
He must also set the price to be paid for each response\footnote{A common guideline for task requesters is to consider \$0.10 per minute to be the minimum wage for  ethical crowdsourcing experiments (\tt{www.wearedynamo.org}).}, which usually includes a participation fee and a bonus based on the quality of the submission and the actual effort required by the task. However, it is a non--trivial problem to set a time limit that gives the workers sufficient time to perform the task correctly without leading to task starvation (i.e., no one working on the task after being assigned).
Generally speaking,  the knowledge of the actual duration of each assignment (task instance) is useful to the requesters for various reasons. First, a task's duration can be used as a proxy to estimate its difficulty, as more difficult tasks usually take longer to complete \cite{faradani2011s}.  Second, this information is useful to set the time limit of a task and to reduce the overall time of task completion. Third. a task requestor can use the task duration to pay fair bonuses to workers based on the difficulty of the task they complete.
When seeking to estimate this information, however, it is important to consider that some workers might not perform a task immediately and they might delay their submissions after accepting the task or, at the other extreme, they might submit a poor annotation in rapid time \cite{kazai2011search}. As a result,  common heuristic  estimates of a task's duration (such as the workers' average or median completion time) that do not account for such aspects are likely to be inaccurate.

Given the above, there are a number of challenges to be addressed in the various steps of designing efficient crowdsourcing workflows. First,  after all the judgments have been collected, the uncertainty about the unknown reliability of individual workers must be taken into account to compute the final labels. Such aggregated labels are often estimated in settings where the true answer of each task is never revealed, as this is the very quantity that the crowdsourcing process is trying to discover \cite{kamar2012combining}. Second, when estimating a  task's duration, the uncertainty over the completion time deriving from the private work schedule of a worker must be taken into account \cite{huff2015these}.
Third, these two challenges must be addressed \emph{simultaneously} due to the interdependencies between the workers' reliability, the time required to complete each task, and the final labels estimated for such tasks.

In an attempt to address these challenges, there has been growing interest in developing algorithms and techniques to compute accurate labels while minimising the set of, possibly unreliable, crowd judgements \cite{sheng2008get}.
In more detail, simple solutions typically use heuristic methods such as majority voting or weighted majority voting \cite{long2013}. However, these methods do not consider the reliability of different workers and they treat all judgments as equally reliable. More sophisticated methods such as the one--coin model \cite{karger2011iterative}, GLAD \cite{whitehill2009whose}, CUBAM \cite{welinder2010multidimensional}, DS \cite{dawid1979maximum} and the {Bayesian Classifier Combination} (BCC) \cite{kim2012bayesian} use probabilistic models that do take reliabilities into account, nor the potential labelling biases of the workers, e.g., the tendency for a worker to consistently over or underrate items. In particular DS represents the worker's skills based on a \emph{confusion matrix} expressing the reliability of a worker for each possible class of objects. BCC works similarly to DS, but it also considers the uncertainty over the confusion matrices and aggregated labels using a principled Bayesian learning framework. This representational power has enabled BCC to be successfully applied to a number of crowdsourcing applications including galaxy classification \cite{simpson2012dynamic}, disaster response \cite{ramchurn2015hac} and sentiment analysis \cite{simpson@language}.
More recently, \cite{venanzi2014community} proposed a community--based extension of BCC (i.e., CBCC) to improve predictions by leveraging groups of workers with similar confusion matrices.
Similarly, \cite{simpson@language} combined BCC with language modelling techniques for automated text sentiment analysis using crowd judgments.
This degree of applicability and performance of BCC-based methods are a promising point of departure for developing new data aggregation methods for crowdsourcing systems. However, none of the existing methods can reason about the workers' completion time to learn the duration of a task outsourced to the crowd. Moreover, all these methods can only learn their probabilistic models from the information contained in the judgment set. Unfortunately, this strategy is challenged by datasets that can be  arbitrarily sparse, i.e., the workers only provide judgments for a small sub-set of tasks, and therefore the judgments only provide weak evidence of the accuracy of a worker. In such contexts, it is our hypothesis that  a wider set of features must be leveraged to learn more reliable crowdsourcing models.
In this work, we focus on the time it takes to a worker to complete a task considered as a key indicator of the quality of his work.
Importantly, the information about the workers' completion time is made available by \emph{all} the most popular crowdsourcing platforms including AMT, the Microsoft Universal Human Relevance System (UHRS) and CrowdFlower. Therefore, we seek to efficiently combine these features into a data aggregation algorithm that can be naturally integrated with the output data  produced by these platforms.
In more detail, we present a novel time--sensitive data aggregation method that simultaneously estimates the tasks' duration  and obtains reliable aggregations of crowdsourced judgments. The characteristic of time--sensitivity of our method relates to its ability to jointly reason about the worker's  completion time together with the judgments in the data aggregation process.
In detail, our method  is an extension of BCC, which we term BCCTime. Specifically, it incorporates a newly developed time model that enables the method to leverage observations of the time spent by a worker on a task to best inform the inference of the final labels.
As in BCC, we use  confusion matrices to represent the labelling accuracy of individual workers. To model the granularity in the workers' time profiles, we use latent variables to represent the propensity of each worker to submit valid judgments.
Further, to model the uncertainty of the duration of each task, we use latent  thresholds to define the time interval within which the task is expected to be completed by all the workers with high propensity to valid labelling.
Then, using Bayesian message-passing inference, our method simultaneously infers the posterior probabilities of (i) the confusion matrix of each worker, (ii) the propensity to valid labelling of each worker,  (iii) the true label of each task and (iv) the upper and lower bound of the duration of each task. In particular, the latter represents a reliable estimate of the likely duration of a task obtained through automatically filtering out all the contributions of the workers with a low propensity to valid labelling.
We demonstrate the efficacy of our method using two commonly--used public datasets that relate to an important Natural Language Processing (NLP) application of crowdsourcing  entity linking tasks. In these datasets,  our method  achieves up to 11\% more accurate classifications compared to seven state-of-the-art methods. Further, we show that our tasks' duration estimates are up to $100\%$ more informative than the common heuristics that do not consider the workers' completion time as correlated to the quality of their judgments.
\newline

Against this background, we make the following contributions to the state of the art.
\begin{itemize}
\item Through an analysis on two real-world datasets for crowdsourcing entity-linking tasks, we show the existence of different types of task--specific quality--time trends, e.g., increasing, decreasing or invariant trends, between the quality of the judgments and the time spent by the workers to produce them. We also re-confirm existing results showing that the workers who submit judgments too quickly or too slowly over the entire task set typically provide lower quality judgments.
\item We develop BCCTime: The first time-sensitive Bayesian aggregation model that leverages observations of a worker's completion time to simultaneously aggregate crowd judgments and  infer the duration of each task as well as the reliability of each worker.
\item We show that BCCTime outperforms seven of the most competitive state--of--the--art data aggregation methods for crowdsourcing, including BCC, CBCC, one coin and majority voting, by providing  up to $11\%$ more accurate classifications and up to $100\%$ more informative estimates of the  task's duration.
\end{itemize}
The rest of the paper unfolds as follows. Section \ref{sec:prel}  describes our notation  and the preliminaries of the Bayesian aggregation of crowd judgments. Section \ref{sec:analysis} details our  time analysis of real-world datasets. Then, Section \ref{sec:model} formally introduces BCCTime and details its probabilistic inference. Section \ref{sec:exp} presents its evaluation against the state of the art.  Section \ref{sec:related} summarises the rest of the related work in the areas of data aggregation and time analysis of crowd generated content and Section \ref{sec:conc} concludes.

\section{Preliminaries}
\label{sec:prel}
Consider a crowd of $K$ workers labelling $N$ objects into $C$ possible classes -- all our symbols are listed in Table \ref{tab:symbols}. Assume that $k$ submits a judgment $c_i^{(k)} \in \{ 1, \dots, C \}$ for classifying an object $i$. Let $t_i$ be the \emph{unobserved} true label of $i$. Then, suppose that $\tau_i^{(k)} \in \mathbb{R}_+$ is the time taken by $k$ to produce $c_i^{(k)}$. Let
$\boldsymbol{J} = \{ c_i^{(k)} | \forall i = 1, \dots, N,  \forall k = 1, \dots, K\}$ and $\boldsymbol{T} = \{ \tau_i^{(k)} | \forall i = 1, \dots, N,  \forall k = 1, \dots, K\}$ be the set containing all the judgments and the time spent by the workers, respectively.
\begin{table}
\caption{List of symbols.}
\centering
\small
\begin{tabular}{ll}
  \hline
  \small Symbol  & \small Definition\\
  \hline
  $N$& Number of tasks\\
  $K$& Number of workers\\
  $C$& Number of true label values\\
    $\boldsymbol{T}$& Set of observed workers' completion time\\
      $\boldsymbol{J}$& Set of observed judgments\\
  $t_i$& True label of the task $i$\\
  $\boldsymbol{t}$& Vector of all $t_i$\\
  $c_i^{(k)}$& Judgment of  $k$ for task $i$\\
   $\tau_i^{(k)}$& Time spent by $k$ for judging the task $i$\\
$\boldsymbol{\pi}^{(k)}$& Confusion matrix of $k$\\
$\boldsymbol{p}$& Class proportions of all the tasks\\
$\psi_k$ & Propensity of $k$ for making valid labelling attempts\\
$\boldsymbol{s}$& Labelling probabilities of a general low-propensity worker \\
  $\boldsymbol{\psi}$& Vector of $\psi_k, \forall k=1, \dots, K$\\
$v_i^{k}$& Boolean variable signalling  if  $c_i^{(k)}$ is a valid labelling attempt\\
$\sigma_i$& Lower-bound threshold of the duration of  task $i$\\
$\lambda_i$& Upper-bound threshold of the duration of task $i$\\
$\sigma_0$& Mean for the Gaussian prior over $\sigma_i$\\
$\gamma_0$& Precision hyperparameter of the Gaussian prior over $\sigma_i$\\
$\lambda_0$& Mean hyperparameter of the Gaussian prior over $\lambda_i$\\
$\delta_0$& Precision hyperparameter of the Gaussian prior over $\lambda_i$\\
$\alpha_0$& True count hyperparameter of the Beta prior over $\psi_k$\\
$\beta_0$& False count hyperparameter of the Beta prior over $\psi_k$\\
$\boldsymbol{s}_0$& Hyperparameter of the Dirichlet prior over $\boldsymbol{s}$\\
$\boldsymbol{p}_0$& Hyperparameter of the Dirichlet prior over $\boldsymbol{p}$\\
$\boldsymbol{\pi}_0^{(k)}$& Hyperparameter of the Dirichlet prior over $\boldsymbol{\pi}^{(k)}$\\
  \hline
\end{tabular}
\label{tab:symbols}
\end{table}

We now introduce the key features of  the BCC model that are relevant to our method. First introduced by \cite{kim2012bayesian}, BCC is a method that combines multiple judgments produced by independent classifiers (i.e., crowd workers) with  unknown accuracy.
Specifically, this model assumes that, for each task $i$, $t_i$ is  drawn from a categorical distribution with parameters $\boldsymbol{p}$:
\begin{align}
 t_i | \boldsymbol{p} \sim \mathrm{Cat}(t_i|\boldsymbol{p})
\end{align}
where $\boldsymbol{p}$ denotes the class proportions for all the objects.
Then, a worker's accuracy is represented through a confusion matrix $\boldsymbol{\pi}^{(k)}$ comprising the labelling probabilities of $k$ for each possible true label value. Specifically, each row of the matrix $\boldsymbol{\pi}_{c}^{(k)} = \{ \pi_{c,1}^{(k)}, \dots, \pi_{c,C}^{(k)} \}$ is the  vector where  $\pi_{c,j}^{(k)}$ is the probability of $k$ producing the judgment $j$ for an object of class $c$.
Importantly, this confusion matrix  expresses both the accuracy (diagonal values) and the biases (off-diagonal values) of a worker. It can  recognise workers who are particularly accurate (inaccurate) or have a bias for a specific class of objects.
In fact,  accurate (inaccurate) workers are represented through high (low) probabilities on the diagonal of the confusion matrix, whilst workers with a bias towards a particular class will have high probabilities on the corresponding column of the matrix.
For example, in the galaxy zoo domain in which the workers classify images of celestial galaxies, the confusion matrices can detect workers who have low accuracy in  classifying spiral galaxies or those who systematically classify every object as elliptical galaxies \cite{simpson2012dynamic}.

To relate the worker's confusion matrix to the quality of a judgment, BCC assumes that $c_{i}^{(k)}$ is drawn from a categorical distribution with parameters corresponding to the $t_i$\emph{-th} row of $\boldsymbol{\pi}^{(k)}$:
\begin{align}
c_{i}^{(k)} | \boldsymbol{\pi}^{(k)}, t_i \sim \mathrm{Cat}\big(c_{i}^{(k)}|\boldsymbol{\pi}_{t_i}^{(k)} \big)
\label{eq:votebcc1}
\end{align}
This is equivalent to having a categorical mixture model over $c_{i}^{(k)}$ with $t_i$ as the mixture parameter and $\boldsymbol{\pi}_{c}$ as the parameter of the $c$-th categorical component.
Then, assuming that the judgments are independent and identically distributed (i.i.d.), the joint likelihood can be expressed as:
\begin{align}
\nonumber p(\boldsymbol{C},  \boldsymbol{t} | \boldsymbol{\pi}, \boldsymbol{p}) = \prod_{i=1}^N \mathrm{Cat}(t_i|\boldsymbol{p}) \prod_{k=1}^K \mathrm{Cat}\big(c_{i}^{(k)}|\boldsymbol{\pi}_{t_i}^{(k)} \big)
\end{align}
Using conjugate Dirichlet prior distributions for the parameters $\boldsymbol{p}$ and $\boldsymbol{\pi}$ and applying Bayes' rule, the joint  posterior distribution can be derived as:
\begin{align}
\nonumber p(\boldsymbol{\pi}, \boldsymbol{p}|\boldsymbol{C}, \boldsymbol{t})\propto& \mathrm{Dir}(\boldsymbol{p}|\boldsymbol{p}_0)\prod_{i=1}^N \nonumber \Big \{ \mathrm{Cat}(t_i|\boldsymbol{p})\\
 &\prod_{k=1}^K \mathrm{Cat}\big(c_{i}^{(k)}|\boldsymbol{\pi}_{t_i}^{(k)} \big) \mathrm{Dir}(\boldsymbol{\pi}_{t_i}^{(k)}|\boldsymbol{\pi}_{t_i, 0}^{(k)} ) \Big \}
\end{align}
From this expression, it is possible to derive the predictive posterior distributions of each unobserved (latent) variable using standard integration rules for Bayesian inference \cite{bishop2006pattern}. Unfortunately, the exact derivation of these posterior distributions is intractable for BCC due to the non-conjugate form of the model \cite{kim2012bayesian}. However, it has been shown that, particularly for BCC models, it is possible to compute efficient approximations of these distributions using standard techniques such as Gibbs sampling \cite{kim2012bayesian}, variational Bayes \cite{simpson2014combined} and Expectation-Propagation  \cite{venanzi2014community}.
Building on this, several extensions of BCC have been proposed for various crowdsourcing domains \cite{venanzi2014community,simpson@language,simpson2012dynamic}. In particular, CBCC applies community--based techniques to represent groups of workers with similar confusion matrices in the classifier combination process \cite{venanzi2014community}. This mechanism enables the model to transfer learning of a worker's reliability through the communities and so improve the quality of the inference.

However, a drawback of all these BCC based models is that they do not learn the task's duration nor do they consider any extra features other than the worker's judgments. As a result, they perform the full learning of the confusion matrices and task labels using only the judgments produced by the workers.
But, as mentioned earlier, this strategy is challenged by sparse datasets where each worker only labels a few tasks.  This is the case, for instance, in the Crowdflower dataset used in the 2013 CrowdScale Shared Task challenge\footnote{\texttt{www.crowdscale.org/shared-task}} where the sentiment of 98,980 tweets was classified by 1,960 workers over five sentiment classes. In this dataset, 30\% of the workers judged only 15 tweets, i.e., 0.015\% of the total samples, and there is a long tail of workers with less than 3 judgments.

\section{Analysis of Workers' Time Spent on Judgments}
\label{sec:analysis}
Having discussed the basic concepts of non-time based data aggregation, we now turn to the analysis of the relationship between the time that workers spend on the task and the quality of the judgments they produce. In contrast to previous works in this area \cite{demartini2012zencrowd,wang2011estimating}, we extend our analysis of quality--time responses to both specific task instances, as well as for the entire task set.
By so doing, we provide key insights to inform the design of our time--sensitive aggregation model. To this end, we consider two public datasets generated from a widely used NLP application of crowdsourcing entity linking tasks.

\subsection{The Datasets}
\label{sec:dataset}
\begin{figure*}[t]
\noindent\begin{minipage}[b]{.45\textwidth}
\centering
\includegraphics[width=7cm]{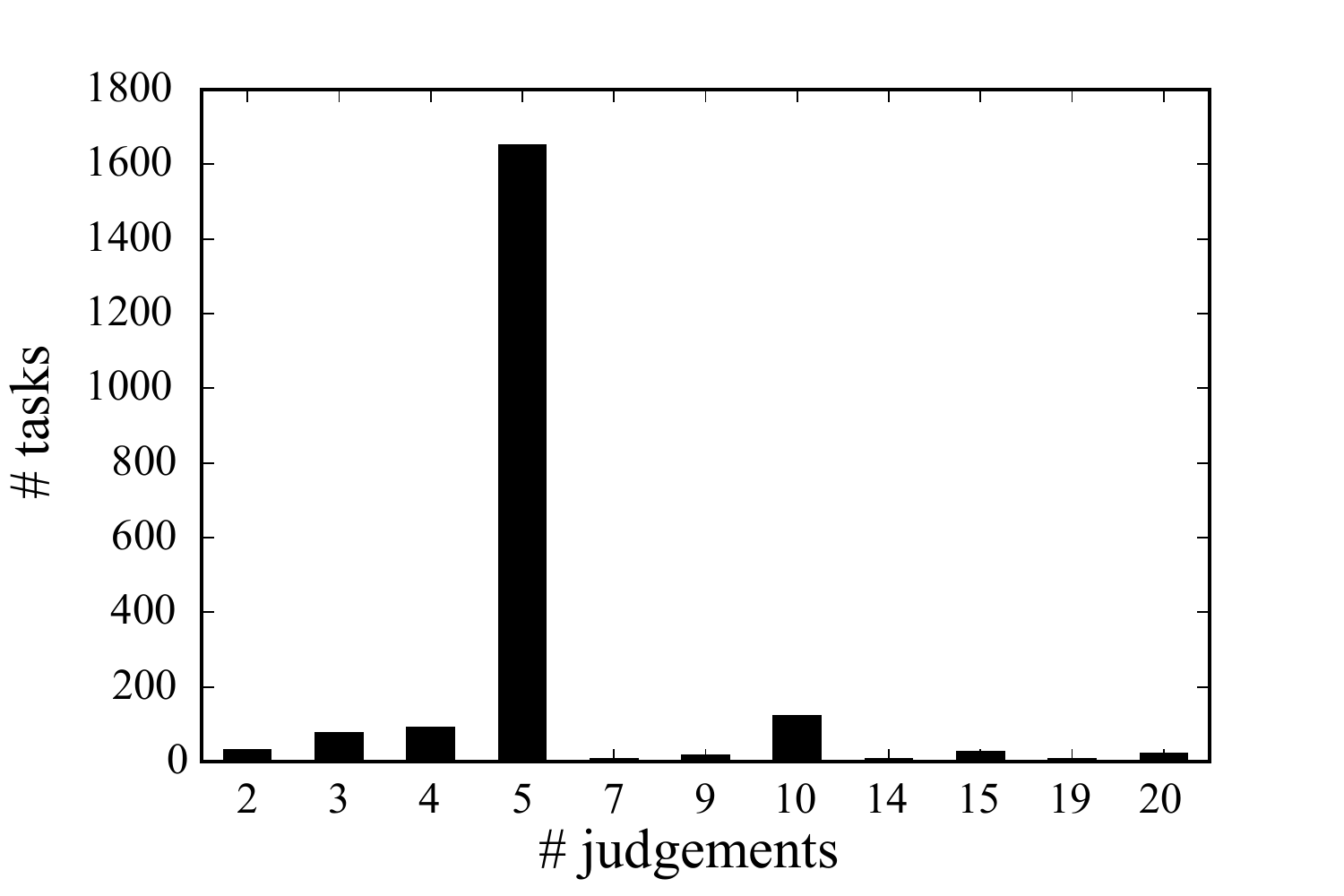}
(a) ZC - IN
\end{minipage}
\noindent\begin{minipage}[b]{.45\textwidth}
\centering
\includegraphics[width=7cm]{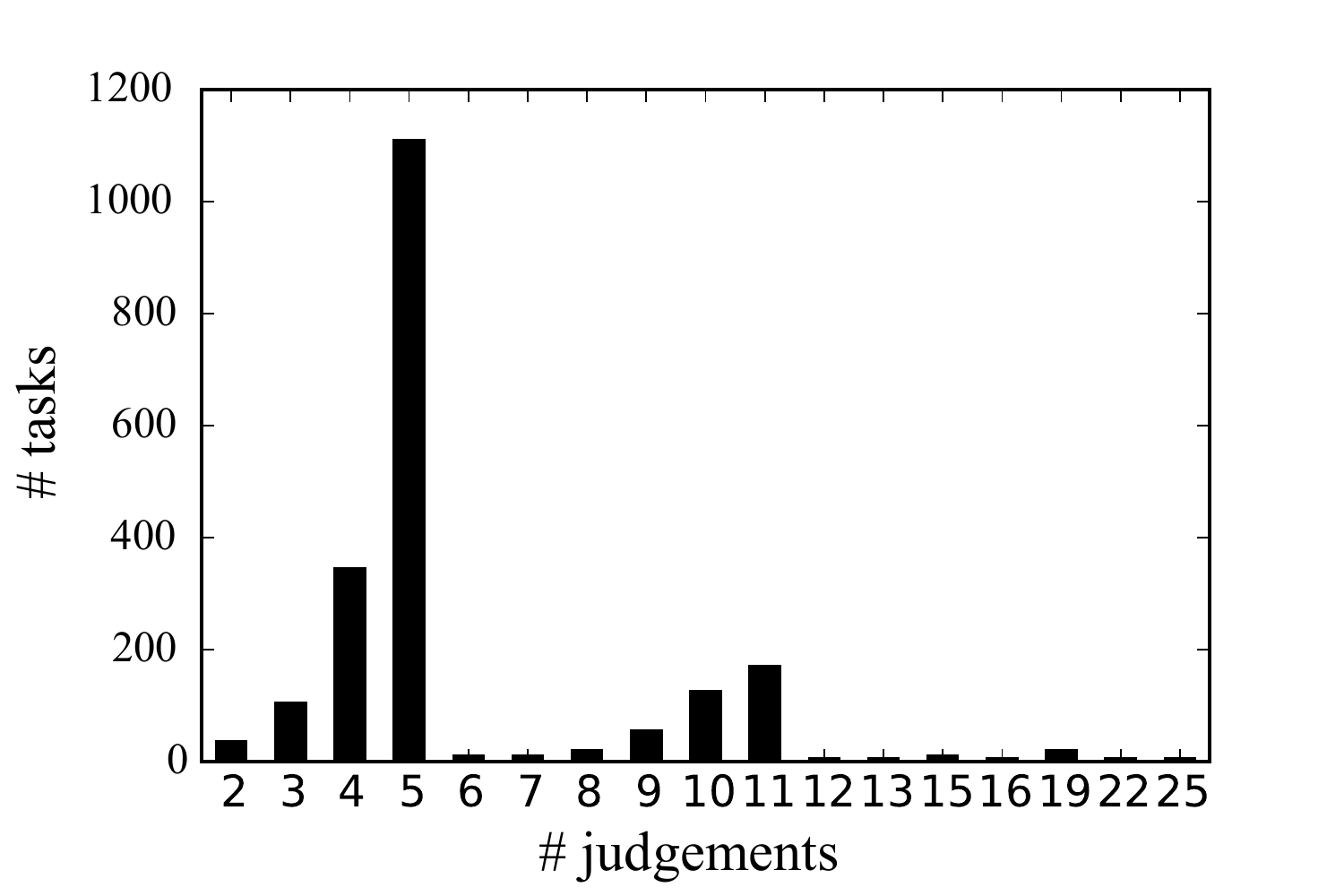}
(b) ZC - US
\end{minipage}
\noindent\begin{minipage}[b]{.45\textwidth}
\centering
\includegraphics[width=7cm]{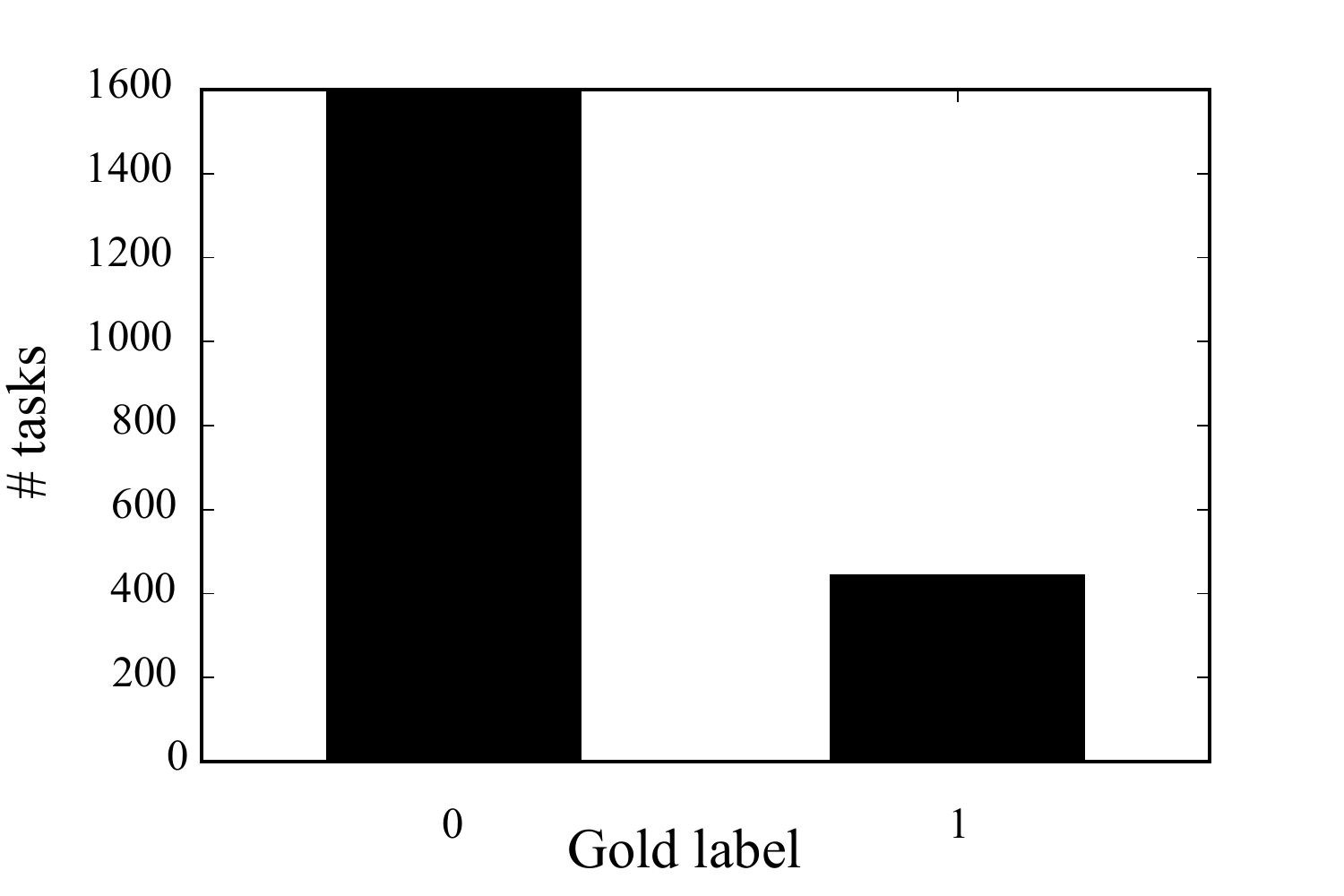}
(c) ZC
\end{minipage}
\hspace{1.2cm}
\noindent\begin{minipage}[b]{.45\textwidth}
\centering
\includegraphics[width=7cm]{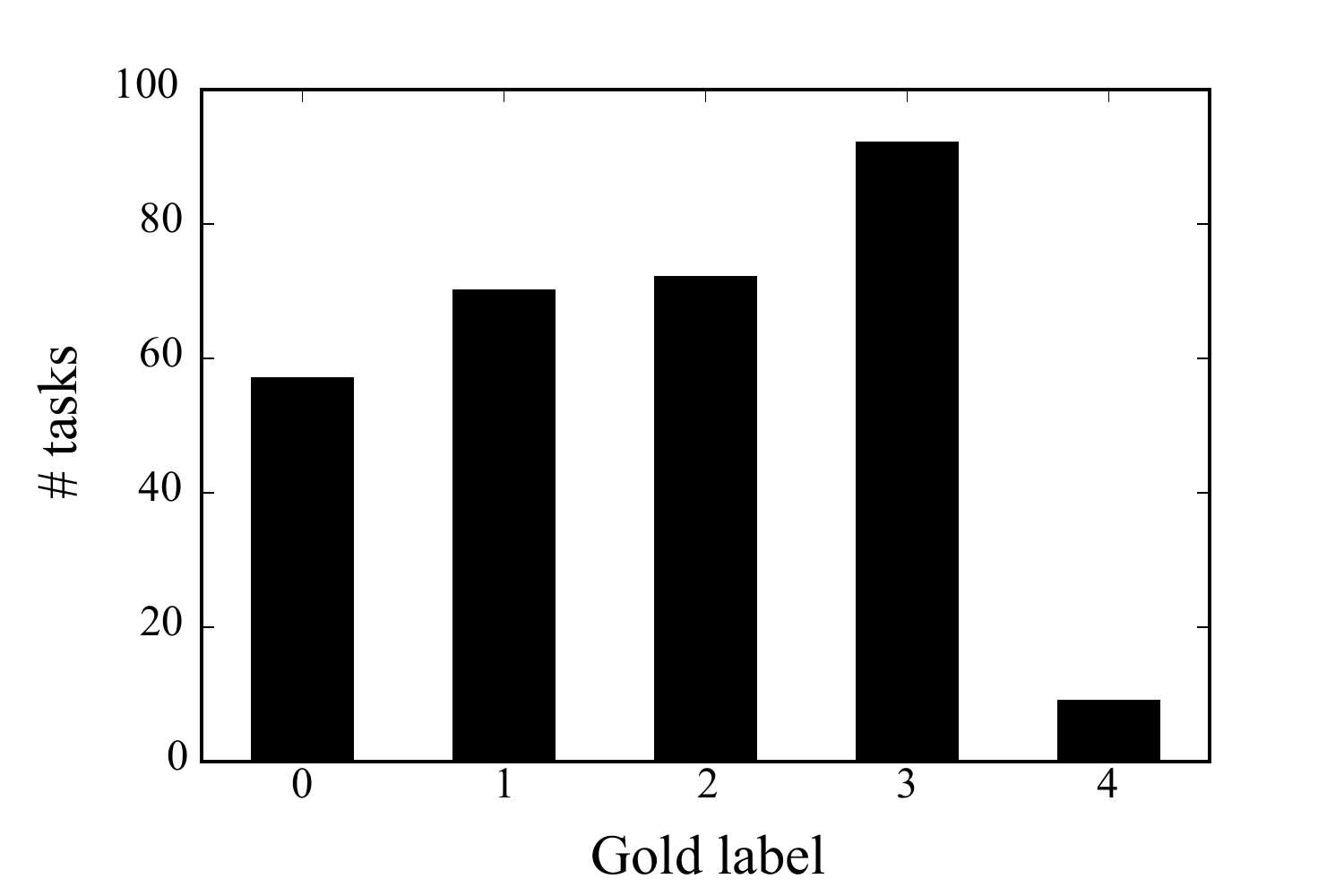}\\(d) WS - AMT
\end{minipage}
\caption{Histograms of the number of judgments per task for ZC-IN (a) and ZC-US (b) -- WS - AMT is not shown because the tasks received exactly 20 judgments -- and the number of tasks per gold label for ZC (c) and WS - AMT (d).}
\label{fig:datasets}
\end{figure*}

\paragraph{ZenCrowd - India (ZC-IN):} contains a set of links between the names of entities extracted from news articles and uniform resource identifiers (URIs) describing the entity  in Freebase\footnote{\tt{www.freebase.com}} and DBpedia\footnote{\tt{www.dbpedia.org}} \cite{demartini2012zencrowd}. The dataset was collected using  AMT, with each worker being asked to classify whether a single URI was either irrelevant (0) or relevant (1) to a single entity. It contains the timestamps of the acceptance and the submission of each judgment. Moreover, gold standard labels were collected from expert editors for all the tasks. No information was released regarding the restrictions on the worker pool, although all workers are known to be living in India, and each worker was paid \$0.01 per judgment. A total of 11,205 judgements were collected from a small pool of 25 workers, giving this dataset a moderately high number of judgements per worker, as detailed in Table \ref{tab:datasets}.
In particular, Figure~\ref{fig:datasets}a shows that the vast majority of tasks receive 5 judgements, while Figure~\ref{fig:datasets}c shows a skewed distribution of gold labels, in which 78\% of links between entities and URIs were classified by workers as {irrelevant} (0). As such, it is worth noting that any binary classifiers with a bias towards unrelated classification will correctly classify the majority of tasks and thus receive a high accuracy. Therefore, as we will detail in Section \ref{sec:exp}, it is important to select accuracy metrics that evaluate the classifier across the whole spectrum of possible discriminant thresholds.

\begin{table*}
\centering
\caption{Crowdsourcing datasets for entity linking tasks.}
\vspace{0.1cm}
\small
\begin{tabular}{lrrrrrrr}
\hline
\hline
 Dataset: & Judgements & Workers & Tasks & Labels & Judgement & Judgements & Judgements \\
  &  &  &  &  & accuracy & per task & per worker \\
  \hline
   ZC-IN &      11205 &      25 &  2040 &      2 &              0.678 &               5.493 &               448.200 \\
   ZC-US &      12190 &      74 &  2040 &      2 &              0.770 &               5.975 &               164.730 \\
   WS-AMT &       6000 &     110 &   300 &      5 &              0.704 &              20.000 &                54.545 \\
   \hline
   \hline
\end{tabular}
\label{tab:datasets}
\end{table*}

\paragraph{ZenCrowd - USA (ZC-US):} This dataset was also provided by \citeauthor{demartini2012zencrowd} (2012) and contains judgements for the same set of tasks as ZC-IN, although the judgements were collected from AMT workers in the US. The same payment of \$0.01 per judgement was used. However, a larger pool of 74 workers was involved, and as such a lower number of judgements were collected from each worker, as shown in Table~\ref{tab:datasets}. Furthermore, Figure~\ref{fig:datasets}b shows a similar distribution of judgements per task as the India dataset, although slightly fewer tasks received 5 judgements, with most of the remaining tasks receiving 3-4 judgements or 9-11 judgements. The judgement accuracy of the US dataset is higher than the India dataset despite an identical crowdsourcing system and reward mechanism being used.

\paragraph{Weather Sentiment - AMT (WS-AMT):} The Weather Sentiment dataset was provided by CrowdFlower for the 2013 Crowdsourcing at Scale shared task challenge.\footnote{\tt{https://www.kaggle.com/c/crowdflower-weather-twitter}} It includes 300 tweets with 1,720 judgements from 461 workers and has been used in several experimental evaluations of crowdsourcing models \cite{simpson@language,venanzi2014community,eps376365}. In detail, the workers were asked to classify the sentiment of tweets with respect to the weather into the following categories: negative (0), neutral (1), positive (2), tweet not related to weather (3) and can't tell (4).  As a result, this dataset pertains to a multi-class classification problem. However, the original dataset used in the Share task challenge did not contain any time information about the collected judgments. Therefore, a new dataset (WS-AMT), was recollected for the same tasks as in the CrowdFlower shared task dataset using the AMT platform, acquiring exactly 20 judgements and recording the elapsed time for each judgment \cite{eps376543}.
As a result, WS-AMT contains 6,000 judgements from 110 workers, as shown in Table~\ref{tab:datasets}. No restrictions were placed on the worker pool and each worker was paid $\$0.03$ per judgement.
Furthermore, Figure~\ref{fig:datasets}d shows that, as per the original dataset, the most common gold label is \emph{unrelated}, while only five tasks were assigned the gold label \emph{can't tell}.

\begin{figure}[t]
\centering
\includegraphics[width=1\textwidth]{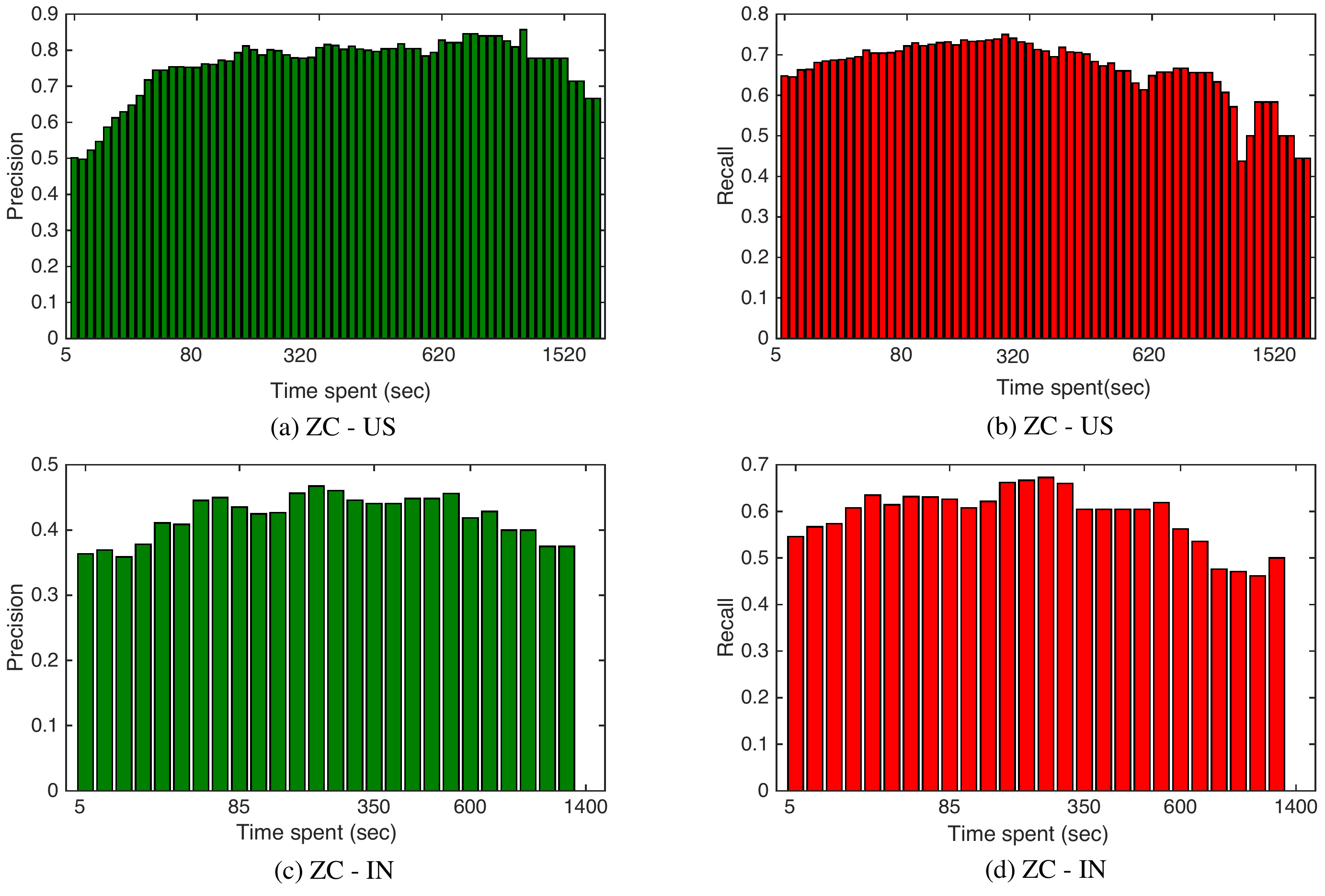}
\vspace{-0.2cm}
\caption{Histograms of the precision and recall binned by the time spent by the US workers (a, b) and Indian workers (c, d) in the ZenCrowd datasets.}
\label{fig:timehist}
\end{figure}

\begin{figure}[ht]
\centering
\includegraphics[width=\textwidth]{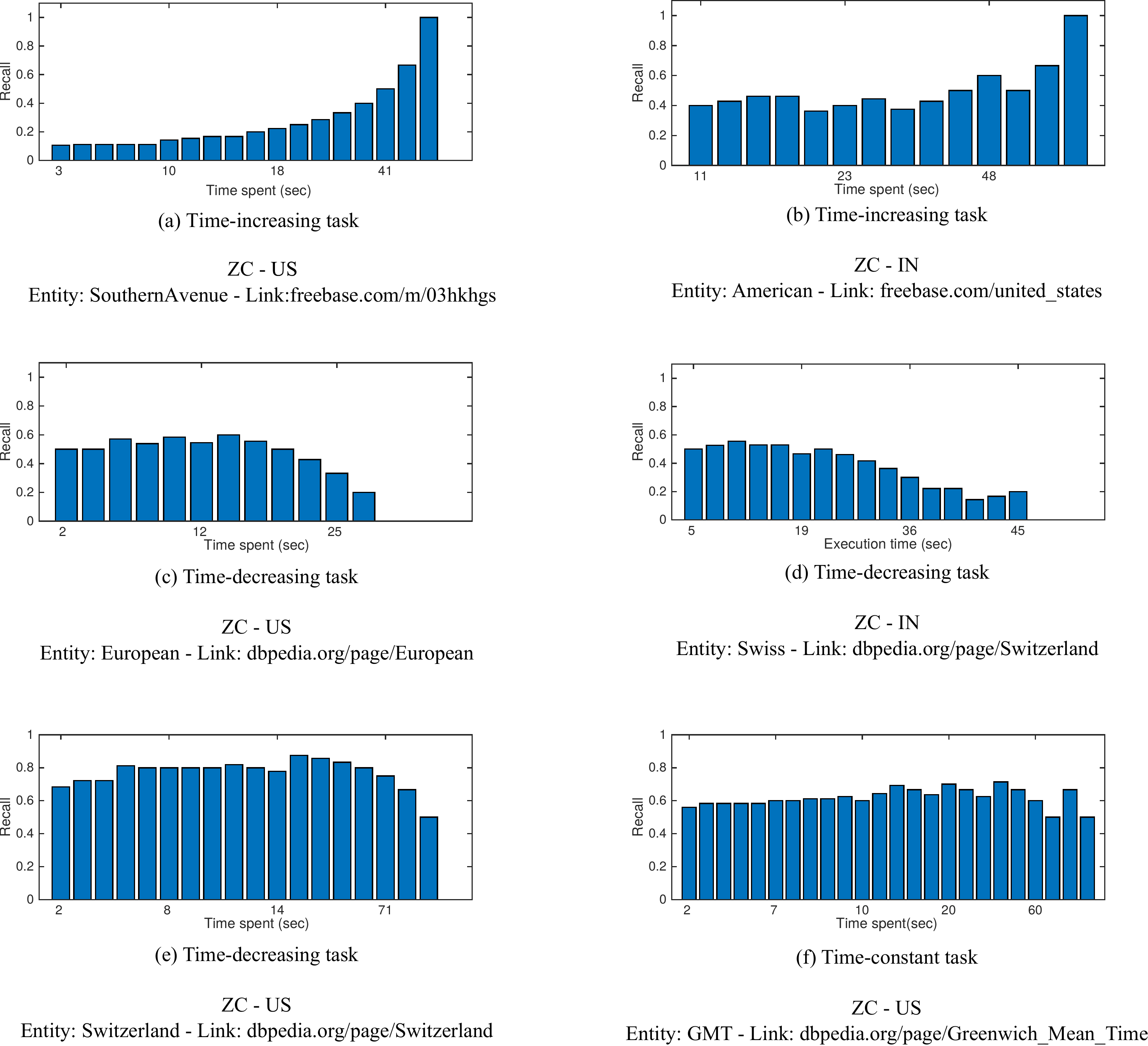}
\vspace{-0.4cm}
\caption{Histograms of the recall for six entity linking tasks with positive gold standard labels and at least ten judgments in the ZenCrowd datasets. They show the different trends of recall-time curves for various tasks.}
\label{fig:tasktimehist}
\end{figure}

\subsection{Time Spent on Task versus Judgment Accuracy}
We wish to analyse the distribution of the workers' completion time and the judgments' accuracy. To do so, we focus on the two datasets, ZC-US and ZC-IN with binary labels. In fact, the binary nature of these two datasets allow us to analyse accuracy at a higher level of detail, i.e., in terms of precision and recall of the workers' judgments and the time spent to produce them. Specifically, Figure \ref{fig:timehist} shows the cumulative distribution of the precision and the recall of the set of judgments selected by a specific time threshold (x-axis) with respect to the gold standard labels.
Here, the precision is the fraction of true positive classifications over all the returned positive classifications (true positives + false positives) and the recall is the number of true positive classifications divided by the number of all the positive samples. Similarly to \citeauthor{demartini2012zencrowd} (2012), we find that the accuracy is lower at the extremes of the time distributions. In ZC-US, both the precision and recall are higher for the sub-set of judgments that were produced in more than 80 seconds and less than 1500 seconds. In ZC-IN, the precision and recall are higher for judgments produced in more than 80 seconds and less than 600 seconds.

In addition, Figure \ref{fig:tasktimehist} shows the distribution of the recall and execution time for a sample set of six {positive} task instances (i.e., entities with positive gold standard labels) with at least ten judgments. For example, Figure \ref{fig:tasktimehist}b shows the time distribution of the judgments for the URI: \texttt{freebase.com/united\_states} associated to the entity ``American''.
In these graphs, some samples have an increasing quality-time curve, i.e., workers spending more time produce better judgments, (Figure \ref{fig:tasktimehist}a and Figure \ref{fig:tasktimehist}b). Other samples have a decreasing quality-time curve, i.e., workers spending more time produce worse judgments (Figure \ref{fig:tasktimehist}c and Figure \ref{fig:tasktimehist}d). Finally, the last two samples have an approximately constant quality-time curve, i.e., worker's quality is invariant to the time spent (Figure \ref{fig:tasktimehist}e and Figure \ref{fig:tasktimehist}f). It can also be seen that these trends naturally correlate to the difficulty of each task instance. For instance, URI: \texttt{freebase.com/m/03hkhgs} linked to the entity ``Southern Avenue"  is  more difficult to judge than the URI: \texttt{dbpedia.org/page/Switzerland} linked to the entity ``Switzerland". In fact, ``Southern Avenue'' is more ambiguous as an entity name, which may lead the worker to open the URI and check its content to be able to issue a correct judgment. Instead, the relevance for the second entity ``Switzerland" can be judged more easily through visual inspection of the URI.
In addition, each task has a specific time interval that includes the sub-set of judgments with the highest precision. For example, in ZC-IN, the judgments with the highest precision for the URI: \texttt{dbpedia.org/page/Switzerland} and the entity ``Switzerland" were submitted between 5 sec. and 20 sec. (Figure \ref{fig:timehist}d). Instead, in ZC-US, the best judgments for the URI: \texttt{dbpedia.org/page/European} linked to the entity ``European" were submitted in the interval of 2 sec. and 16 sec. (Figure \ref{fig:timehist}c).
As a result, it is clear that each task instance has specific quality--time profile that relates to the difficulty of labelling that instance.

\begin{figure}[t]
\centering
\includegraphics[width=1.02\textwidth]{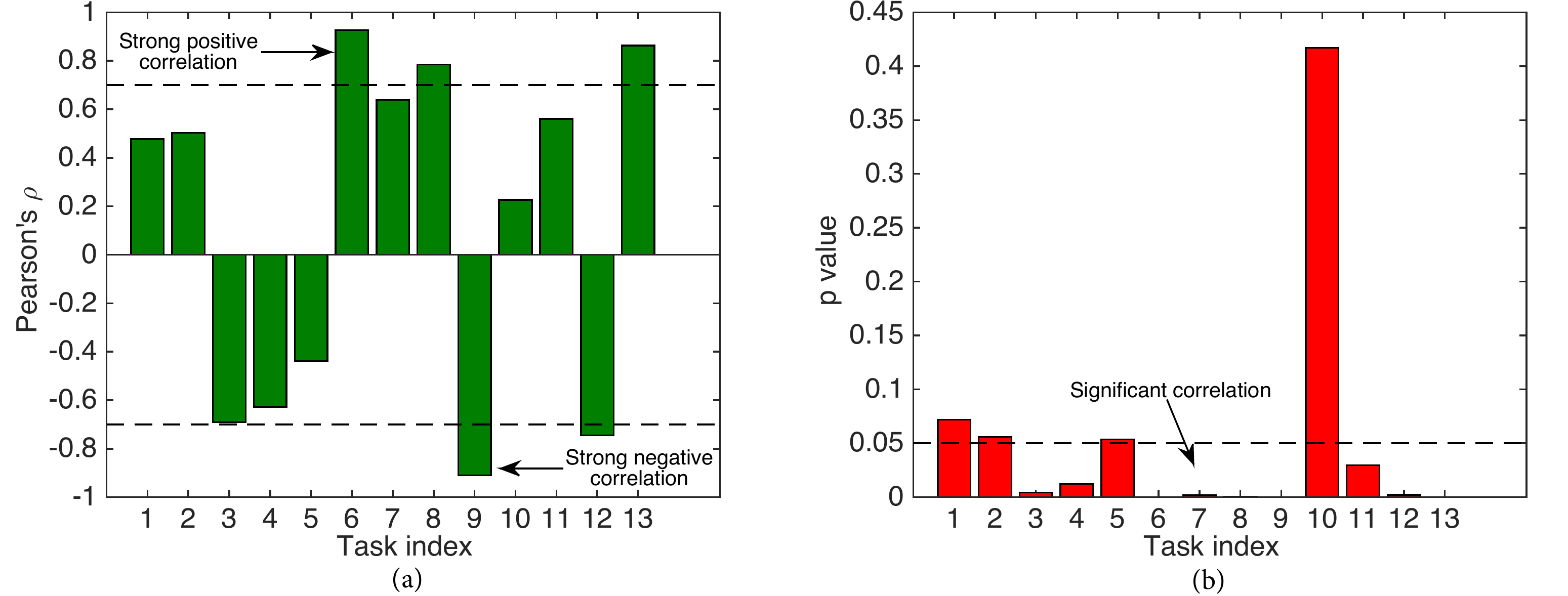}
\vspace{-0.4cm}
\caption{The Pearson's correlation coefficient (a) and the p-value (b) of the linear correlation between the workers' completion time and the judgments accuracy for the 13 entity linking tasks with positive gold standard labels and more than the judgments in the ZenCrowd datasets.}
\label{fig:pearson}
\end{figure}

To better analyse these trends, Figure \ref{fig:pearson} shows the Pearson's  correlation coefficient ($\rho$) (i.e., a standard measure of the degree of linear correlation between two variables) for all the 13 entities with positive links and more than ten judgments across the two datasets. The time spent by the worker is not always (linearly) correlated to the quality of the judgment across all the task instances. Some tasks have a significantly positive correlation (i.e., task index = 6, 8, 13 with $\rho>0.7$, $p < 0.05$), others have a significantly negative correlation (i.e., task index = 9, 12 with $\rho<0.7$, $p < 0.05$),  whilst the other tasks have a less significant correlation between the accuracy of their judgments and the time spent by the workers.
This confirms that different task instances have substantially different quality-time responses based on the difficulty of each sample. Thus, this insight significantly extends the previous findings reported by \citeauthor{demartini2012zencrowd} (2013) in which such a quality--time trend was only observed across the entire task set.
Moreover, it empirically supports the theory of several existing data aggregation models \cite{kamar2015identifying,whitehill2009whose,bachrach2012grade} that make use of these task--specific features to achieve more accurate classifications in a number of crowdsourcing applications concerning, among others, galaxy classification \cite{kamar2015identifying}, image labelling \cite{whitehill2009whose} and problem solving \cite{bachrach2012grade}.

\section{The BCCTime Model}
\label{sec:model}
Based on the above results of the time analysis of workers' judgments, we observed that different types of quality--time trends occur for specific task instances. However, the standard BCC, as well as all the other existing aggregation models  that do not consider this information, are unable to perform inference over the likely duration of a task.
To rectify this, there is a need to extend BCC to be able to include these  trends in the aggregation of crowd judgments. To this end, the model must be flexible enough to identify workers who, in addition to having imperfect skills, may also not have the intention to make a valid attempt to complete a task. This further increases the uncertainty about data reliability. In this section, we describe our Bayesian Classifier Combination model with Time (BCCTime). In particular, we describe the  three components of the model concerning (i) the representation of the unknown workers' propensity to valid labelling, (ii) the reliability of workers' judgments and (iii) the uncertainty in the worker's completion time, followed by the details of its probabilistic inference.

\subsection{Modelling Workers' Propensity To Valid Labelling}
Given the uncertainty about the intention of a worker to submit valid judgments, we introduce the latent variable $\psi_k \in [0, 1]$ representing the {propensity} of $k$ towards making a valid labelling attempt for any given task. In this way, the model is able to naturally explain the unreliability of a worker based not only on her imperfect skills but also on her attitude towards approaching a task correctly.
In particular, $\psi_k$  close to one means that the worker has a tendency to exert her best effort to provide valid judgments, even though her judgments might be still noisy as a consequence of the imperfect skills she possesses. In contrast, $\psi_k$ close to zero means that the worker tends to not provide valid judgments for her tasks, which means that she behaves similarly to a spammer.
Specifically, only the workers with high propensity to valid labelling will provide inputs that are meaningful to the task's true label and the task's duration. To capture this, we define a per-judgment boolean variable $v_i^{(k)} \in \{0, 1\}$ with $v_i^{(k)}=1$ meaning that $k$ has made a valid labelling attempt when submitting $c_{i}^{(k)}$ and $v_i^{(k)}=0$ meaning that $c_{i} ^{(k)}$ is an invalid annotation. In this setting, the number of valid labelling attempts made by the worker derives from her propensity to valid labelling. Thus, we can model this by assuming that $v_i^{(k)}$ is a random draw from a Bernoulli distribution parametrised by $\psi_k$  :
\begin{align}
v_i^{(k)} \sim \mathrm{Bernoulli}(\psi_k)
\end{align}

\noindent That is, workers with high propensity to valid labelling are more likely to make more valid labelling attempts, whilst workers with  low propensity are more likely to submit more spam annotations.

\subsection{Modelling Workers' Judgments}
\label{sec:jud}
Here we describe the part of the model concerned with the generative process of crowd judgments from the confusion matrix and the propensity of the workers.
Intuitively, only those judgments associated with valid labelling attempts should be considered to estimate the final labels.
This means that each judgment may be generated from two different processes depending on whether or not it comes from a valid labelling attempt.
To capture this in the generative model of BCCTime, a mixture model is used to switch between these two cases conditioned on $v_i^{(k)}$. For the first case of a valid labelling attempt, i.e., $v_i^{(k)}=1$,  the judgment is generated through the worker's confusion matrix as per the standard BCC model. Therefore, we assume that $c_{i}^{(k)}$ is generated  for the same model described for BCC (Eq. \ref{eq:votebcc1}), including $v_i^{(k)}$ in the conditional variables. Formally:
\begin{align}
c_{i}^{(k)} | \boldsymbol{\pi}^{(k)}, t_i, v_i^{(k)}=1 \sim \mathrm{Cat}\big(c_{i}^{(k)}|\boldsymbol{\pi}_{t_i}^{(k)} \big)
\label{eq:vote1}
\end{align}
For  the second case of  a judgment produced from an invalid labelling attempt, i.e., $v_i^{(k)}=0$,  it is natural to assume that the judgment does not contribute to the estimation of the true label. Formally, this assumption can be represented through general random vote model in which $c_{i}^{(k)}$ is drawn from a categorical distribution with a vector parameter $\boldsymbol{s}$:
\begin{align}
c_{i}^{(k)} | \boldsymbol{s}, v_i^{(k)}=0 \sim \mathrm{Cat}\big(c_{i}^{(k)}|\boldsymbol{s} \big)
\label{eq:vote2}
\end{align}
Here $\boldsymbol{s}$ is the vector of the labelling probabilities of a general worker with low propensity to make valid labelling attempts. Notice that the equation above does not depend on $t_i$, which means that all the judgments coming from invalid labelling attempts are treated as noisy responses uncorrelated to $t_i$.

\subsection{Modelling Workers' Completion Time}
\label{sec:time}
 As shown in Section \ref{sec:analysis}, the duration of a task may be defined as the interval in which workers are more likely to submit high-quality judgments. However, due to the dependency of the duration on the task's characteristics, the requirement is that such an interval must be non-constant across all the tasks.
To model this, we define a lower-bound threshold, $\sigma_i$,  and an upper-bound threshold, $\lambda_i$, for the time interval representing the duration of $i$. Both these per--task thresholds are latent variables that must be learnt at training time.
Then, the tasks with a lower or higher variability in their duration can be represented based on the values of their time thresholds.
 In this setting, all the valid labelling attempts made by the workers are expected to be completed within the task's duration interval  detailed by these thresholds.
 Formally, we represent the probability of $\tau_i^{(k)}$ being greater than $\sigma_i$ using the standard \emph{greaterThan} probabilistic factor introduced by \cite{export:67956} for the TrueSkill Bayesian ranking model:

\begin{align}
 \mathbb{I}(\tau_i^{(k)} > \sigma_i | v_i^{(k)}=1)
 \label{eq:factor1}
\end{align}

\noindent This factor defines a non-conjugate relationship over $\sigma_i$ such that  the posterior distribution of $\tau_i^{(k)}$ is not in the same form as the prior distribution of $\sigma_i$. Therefore the posterior distribution ${{p}}(\tau_i^{(k)})$ needs to be approximated. We do this via moment matching with a Gaussian distribution $\hat{p}(\tau_i^{(k)})$ by matching the precision and the precision adjusted mean (i.e., the mean multiplied by the precision) to the posterior distribution of ${p}(\tau_i^{(k)})$, as shown in Table 1 in \citeauthor{export:67956} (2007).
In a similar way, we model the probability of $\tau_i^{(k)}$ being greater than $\lambda_i$ as:

\begin{align}
 \mathbb{I}( \lambda_i > \tau_i^{(k)} | v_i^{(k)}=1)
  \label{eq:factor2}
\end{align}

\noindent Drawing all this together, upon observing a set of i.i.d. pairs of judgments and workers' completion times contained in $\boldsymbol{J}$ and $\boldsymbol{T}$ respectively, we can express the joint likelihood of BCCTime as:

\begin{align}
\nonumber \quad p(\boldsymbol{J}, \boldsymbol{T}, \boldsymbol{t} | \boldsymbol{\pi}, \boldsymbol{p}, \boldsymbol{s}, \boldsymbol{\psi}) =& \prod_{i=1}^N \mathrm{Cat}(t_i|\boldsymbol{p}) \bigg \{ \prod_{k=1}^K \Big( \mathbb{I}(\tau_i^{(k)} > \sigma_i ) \mathbb{I}(\lambda_i > \tau_i^{(k)}) \mathrm{Cat}\big(c_{i}^{(k)}|\boldsymbol{\pi}_{t_i}^{(k)} \big)\Big)^{\psi_k}\\
&  \mathrm{Cat}\big(c_{i}^{(k)}|\boldsymbol{s} \big) ^{(1 - \psi_k)} \bigg \}
\end{align}

\begin{figure*}
\centering
\def\svgwidth{\textwidth}
{\small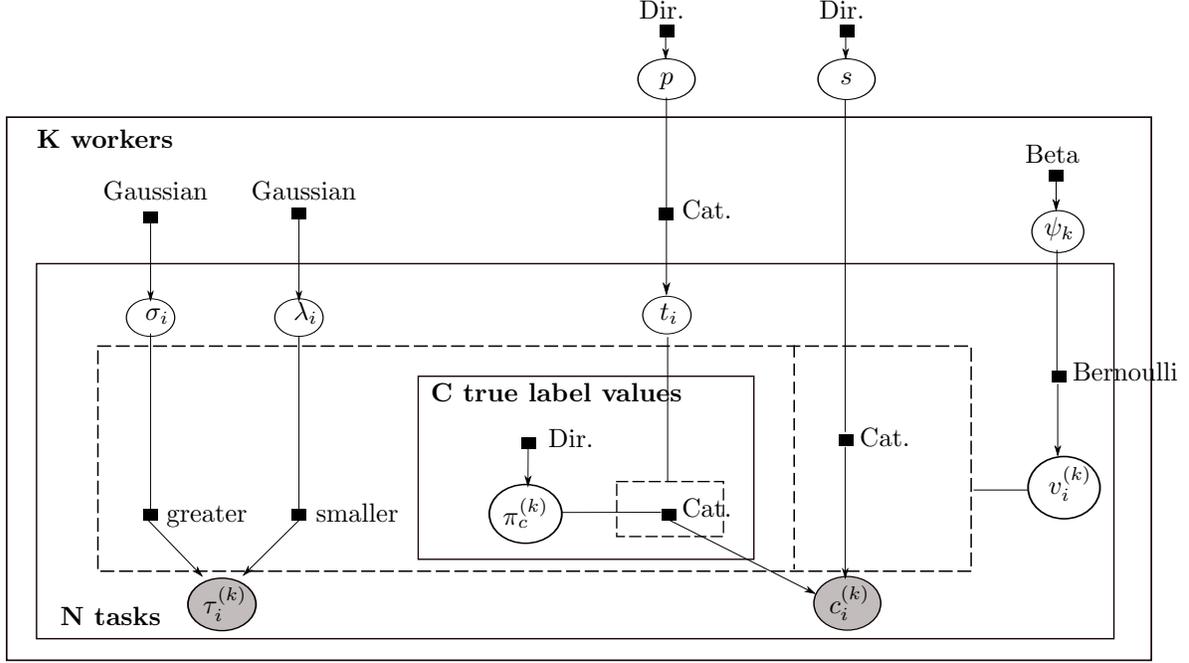}
\caption{The factor graph of BCCTime.}
\label{fig:factor}
\end{figure*}

\noindent The factor graph of BCCTime is illustrated in Figure \ref{fig:factor}. Specifically, the two shaded variables $c_i^{(k)}$ and $\tau_i^{(k)}$ are the observed inputs, while all the unobserved random variables are unshaded. The graph uses the \emph{gate} notation (dashed box) introduced by \cite{minka2008gates} to represent the two mixture models of BCCTime. Specifically, the outer gate represents the workers' judgments (see Section \ref{sec:jud}) and completion times (see Section \ref{sec:time}) that are generated from either BCC or the random vote model using $v_i^{(k)}$ as the gating variable. The inner gate is the mixture model generating the workers' judgments from the rows of the confusion matrix using $t_i$ as the gating variable.

\subsection{Probabilistic Inference}
To perform Bayesian inference over all the unknown quantities, we must provide prior distributions for the latent parameters of BCCTime. Following the structure of the model,  we can select conjugate distributions for all such parameters to enable a more tractable inference of their posterior probabilities. Therefore, the prior of $\boldsymbol{p}$ is Dirichlet distributed with hyperparameter $\boldsymbol{p}_0$:
\begin{align}
\text{(true label prior)} \quad \boldsymbol{p}&\sim \mathrm{Dir}(\boldsymbol{p} | \boldsymbol{p}_0)
\end{align}

\noindent The priors of $\boldsymbol{s}$ and $\boldsymbol{\pi}_c^{(k)}$ are also Dirichlet distributed with hyperparameter $\boldsymbol{s}_0$ and $\boldsymbol{\pi}_{c,0}^{(k)}$ respectively:
\begin{align}
\text{(spammer label prior)} &\quad\boldsymbol{s}\sim \mathrm{Dir}(\boldsymbol{s}|\boldsymbol{s}_0)\\
\text{(confusion matrix prior)} &\quad{\boldsymbol{\pi}_c^{(k)}}\sim \mathrm{Dir}(\boldsymbol{\pi}_c^{(k)}|\boldsymbol{\pi}_{c,0}^{(k)})
\end{align}
Then, $\psi_k$ has a Beta prior with true count $\alpha_0$ and false count $\beta_0$:
\begin{align}
\text{(worker's propensity prior)} \quad \psi_k&\sim \mathrm{Beta}(\psi_k | \alpha_{0}, \beta_{0})
\end{align}
The two time thresholds $\sigma_i$ and $\lambda_i$ have Gaussian priors with mean $ \sigma_0$ and $\lambda_0$ and precision  $\gamma_0$ and $ \delta_0 $ respectively:
\begin{align}
\text{(lower-bound of the task's duration threshold prior)} \quad\sigma_i&\sim \mathcal{N}(\sigma_i | \sigma_{0}, \gamma_{0})\\
\text{(upper-bound of the task's duration threshold prior)} \quad\lambda_i&\sim \mathcal{N}(\lambda_i | \lambda_{0}, \delta_{0})
\end{align}
Collecting all the hyperparameters in the set $\boldsymbol{\Theta} = \{ \boldsymbol{p}_0, \boldsymbol{s}_0, \alpha_0, \beta_0, \sigma_0, \gamma_0, \lambda_0, \delta_0 \}$, we find by applying Bayes' theorem that the joint posterior distribution is proportional to:

\begin{align}
\quad \nonumber p(\boldsymbol{\pi}, \boldsymbol{p}, \boldsymbol{s}, \boldsymbol{t} ,\boldsymbol{\psi} | \boldsymbol{J}, \boldsymbol{T}, \boldsymbol{\Theta})  \propto &~\mathrm{Dir}(\boldsymbol{s}|\boldsymbol{s}_0)\mathrm{Dir}(\boldsymbol{p}|\boldsymbol{p}_0)\prod_{i=1}^N \bigg\{ \mathrm{Cat}(t_i|\boldsymbol{p})  \mathcal{N}(\sigma_i | \sigma_{0}, \gamma_{0})\mathcal{N}(\lambda_i | \lambda_{0}, \delta_{0}) \\
\nonumber& \prod_{k=1}^K  \Big( \mathbb{I}(\tau_i^{(k)} > \sigma_i )\mathbb{I}( \lambda_i < \tau_i^{(k)} ) \mathrm{Cat}\big(c_{i}^{(k)}|\boldsymbol{\pi}_{t_i}^{(k)} \big) \mathrm{Dir}(\boldsymbol{\pi}_{t_i}^{(k)}|\boldsymbol{\pi}_{t_i,0}^{(k)}) \Big)^{\psi_k} \\
&\mathrm{Cat}\big(c_{i}^{(k)}|\boldsymbol{s} \big)^{(1 - \psi_k)} \mathrm{Beta}(\psi_k | \alpha_{0}, \beta_{0}) \bigg\}
\end{align}

\noindent From this expression, we can compute the marginal posterior distributions of each latent variable by integrating out all the remaining variables. Unfortunately, these integrations are intractable due to the non--conjugate form of our model. However, we can still compute approximations of such posterior distributions using standard techniques from the family of approximate Bayesian inference methods \cite{minka2001family}. In particular, we use the well-known EP algorithm \cite{minka2001expectation} that has been shown to provide good quality approximations for BCC models \cite{venanzi2014community}\footnote{Alternative inference methods such as Gibbs sampling or Variational Bayes can be trivially applied to our model in the Infer.NET framework.}. This method leverages a factorised distribution of the joint probability to approximate the marginal posterior distributions through an iterative message passing scheme implemented on the factor graph. Specifically, we use the EP implementation provided by Infer.NET \cite{minkainfer}, which is a standard framework for running Bayesian inference in probabilistic models. Using Infer.NET, we are able to train BCCTime on our largest dataset of 12,190 judgments within seconds using approximately 80MB of RAM on a standard laptop.

\section{Experimental Evaluation}
\label{sec:exp}
Having described our model, we test its performance in terms of classification accuracy and ability to learn the tasks' duration in real crowdsourcing experiments. Using the datasets described in Section \ref{sec:analysis}, we conduct experiments in the following experimental setup.

\subsection{Benchmarks}
We consider a set of benchmarks consisting of three popular baselines (Majority voting, Vote distribution and Random) and three state--of--the--art aggregation methods (One coin, BCC and CBCC) that are commonly employed in crowdsourcing applications. In more detail:

\begin{itemize}
\item \emph{One coin:} This method represents the accuracy of a worker with a single reliability parameter (or worker's coin) assuming that the worker will return the correct answer with probability specified by the coin, and the incorrect answer with inverse probability. As a result, this method is only applicable to binary datasets.
Crucially, this model represents the core mechanism of several existing methods including \cite{whitehill2009whose,demartini2012zencrowd,liu2012variational,karger2011iterative,li2014wisdom}\footnote{In particular, we refer to \emph{One coin} as the unconstrained version of \emph{ZenCrowd} \cite{demartini2012zencrowd} without the two \emph{unicity} and \emph{SameAs} constraints defined in the original method. This suggests that this version is more suitable for a fair comparison with the other methods.}. 
\item \emph{BCC:} This is the closest benchmark to our method that was described in Section \ref{sec:prel}. It learns the confusion matrices and the aggregated labels without considering the worker's completion time as an input feature. It has been used in several crowdsourcing contexts including galaxy classification \cite{simpson2012dynamic}, image annotation \cite{kim2012bayesian} and disaster response \cite{ramchurn2015hac}.
\item \emph{BCCPropensity:} This is equivalent to BCCTime where only the workers' propensity is learnt. This benchmark is used to assess the contribution of inferring only the worker's propensity, versus their joint learning with the tasks' time thresholds, to the quality of the final labels. Note that BCCPropensity is easy to obtain from BCCTime by setting the time thresholds to static observations  with $\sigma = 0.0$ and $\lambda = \mathrm{max. value}$.
\item \emph{CBCC:} An extension of BCC that learns the communities of workers with similar confusion matrices as described in Section \ref{sec:prel}.
 Given a judgment set, CBCC is able to learn  the confusion matrix of each community and each worker, as well as the task label. This method has also been used in a number of crowdsourcing applications including web search evaluation and sentiment analysis \cite{venanzi2014community}. In our experiments, we ran CBCC with the number of worker types set to two communities in order to infer the two groups of more reliable workers and less reliable workers -- similar results were observed for higher number of communities.
\item \emph{Majority Voting:} This is a simple yet very popular algorithm that estimates the aggregated label as the one that receives the most votes \cite{littlestone_weighted_2002,long2013}. It assigns a point mass to the label with the highest consensus among a set of judgments. Thus, the algorithm does not represent its uncertainty around a classification and it considers all judgments as coming from reliable workers.
\item \emph{Vote Distribution:} This method estimates the true label based on the empirical probabilities of each class observed in the judgment set \cite{simpson@language}. Specifically, it assigns the probability of a label as the fraction of judgments corresponding to that label.
\item \emph{Random:} This is a baseline method that assigns random class labels to all the tasks, i.e., it assigns uniform probabilities to all the labels.
\end{itemize}
Note that the alternative variant of BCCTime that captures only the time spent is redundant. In fact, when the workers' propensity is not modelled together with the time spent, the workers' accuracy is only captured by their confusion matrices. This means that the model is equivalent to BCC, which is already included in the benchmarks. All these benchmarks were also implemented in Infer.NET and trained using the EP algorithm.
In our experiments, we set the hyperparameters of BCCTime to reproduce the typical situation in which the task requester has no prior knowledge of the true labels and the labelling probabilities of the workers, and only a basic prior knowledge about the accuracy of workers representing that, a priori, they are assumed to be better than random annotators \cite{kim2012bayesian}.  Therefore, the workers' confusion matrices are initialised with a slightly higher value on the diagonal (0.6) and lower values on the rest of the matrix. Then, the Dirichlet priors for $\boldsymbol{p}$ and $\boldsymbol{s}$ are set uninformatively with uniform counts\footnote{It should be noted, however, that in cases where a different type of knowledge is available about the workers, this information can be plugged into our method  by selecting the appropriate prior distributions.}. The priors of the confusion matrices were initialised with a higher diagonal value (0.7) meaning that a priori the workers are assumed to be better than random. The Gaussian priors for the tasks' time durations are set with means $\sigma_0 = 10$ and $\lambda_0=50$ and precisions $\gamma_0 = \delta_0 = 10^{-1}$, meaning that a priori each entity linking task is expected to be completed within $10$ and $50$ seconds. Furthermore, we initialise the Beta prior of $\psi_k$ as a function of the number of tasks with $\alpha_0 = 0.7N$ and $\beta_0 = 0.3N$ to represent the fact that a priori the worker is considered as a reliable if she makes valid labelling attempts for $70\%$ of the tasks. Importantly, given the shape distribution of the worker's time completion data observed in the datasets (see Figure \ref{fig:timehist}), we apply a logarithmic transformation to $\tau_i^{(k)}$ in order to obtain a  more  uniform distribution of workers' completion time in the training data.
Finally, the priors of all the benchmarks were set equivalently to BCCTime.

\subsection{Accuracy Metrics}
We evaluate the classification accuracy of the tested methods as measured by the Area Under the ROC Curve (AUC) for ZC-US and ZC-IN and the average recall for WS-AMT. In particular, the former is a standard accuracy metric to evaluate the performance of binary classifiers over a range of discriminant thresholds applied to their predictive class probabilities \cite{hanley1982meaning}, which is well suited for the two ZenCrowd binary datasets. The latter is the recall averaged over the class categories \cite{rosenberg2012classifying}, which is the main metric used to score the probabilistic methods that competed in the 2013 CrowdFlower shared task challenge on a dataset equivalent to WS-AMT (see Section \ref{sec:dataset}).

\begin{table}[h]
    \begin{minipage}[t]{0.45\textwidth}
        \centering
        \caption{The \emph{AUC} of the tested methods measured on the ZenCrowd datasets. The highest AUC in each dataset is highlighted in bold.}
        \begin{tabular}{lcc}
            \hline
            \hline
             Dataset: & ZC-US & ZC-IN \\
             \hline
             Majority vote &0.3820&0.3862\\
             Vote distribution &0.2101&0.3080\\
             One coin & 0.7204 &0.6263\\
             Random & 0.5000&0.5000\\
             BCC & 0.6418&0.5407\\
             CBCC & 0.6730&0.5544\\
             BCCPropensity &0.7740 & 0.6177\\
             BCCTime & \textbf{0.7800}&\textbf{0.6925}\\
                \hline
                \hline
        \end{tabular}
        \label{tab:auc}
    \end{minipage}
    \hfill
\begin{minipage}[t]{0.45\textwidth}
\centering
\caption{The \emph{average recall} of the tested methods measured on WS-AMT. The highest average recall is highlighted in bold.}
\centering
\begin{tabular}{lc}
\hline
\hline
 Dataset: & WS-AMT \\
 \hline
 Majority vote &0.727\\
 Vote distribution &0.728\\
 One coin &N/A\\
 Random &0.183\\
 BCC &0.705\\
  CBCC &0.711\\
 BCCPropensity &0.703\\
 BCCTime &\textbf{0.730}\\
   \hline
   \hline
\end{tabular}
\label{tab:avg_rec}
\end{minipage}
\end{table}
\begin{table*}[b]
\centering
\caption{The propensity of workers learnt from BCCTime in each dataset.}
\vspace{0.1cm}
\begin{tabular}{lcc}
\hline
\hline
 Dataset: & \multicolumn{1}{l}{\% high propensity} & \multicolumn{1}{l}{\% low propensity}  \\
 &workers ($p(\psi_k)>0.5$)&workers ($p(\psi_k)\leq0.5$)\\
  \hline
   ZC-US &      93.2\%  & 6.8\% \\
   ZC-IN &     60\%  &  30\% \\
   WS-AMT & 97.3\% & 2.7\% \\
   \hline
   \hline
\end{tabular}
\label{tab:prop}
\end{table*}

\subsection{Results}
Table \ref{tab:auc} reports the AUC of the seven algorithms on the ZenCrowd datasets. Specifically, it shows that BCCTime and BCCPropensity have the highest accuracy in both the datasets: Their AUC is $11\%$ higher in ZC-IN and $8\%$ higher in ZC-US, respectively,  compared to the other methods. Among the two, BCCTime is the best method with an improvement of $13\%$ in ZC-IN and $1\%$ in ZC-US. Similarly, Table \ref{tab:avg_rec} reports the average recall of the methods in WS-AMT showing that BCCTime has the highest average recall,  which is $2\%$ higher than the second best benchmark (Vote distribution) and $4\%$ higher than BCCPropensity. This means that the inference of the time thresholds, which already provides valuable information about the tasks extracted from the judgments, also adds an extra quality improvement to aggregated labels in addition to the modelling of the workers' propensities. 
This is an important observation because it proves that the information of workers' completion time can be effectively for data aggregation.
Altogether, this information allows the model to correctly filter unreliable judgments and consequently provide more accurate classifications. 

\begin{figure*}[t]
\centering
\includegraphics[width=\textwidth]{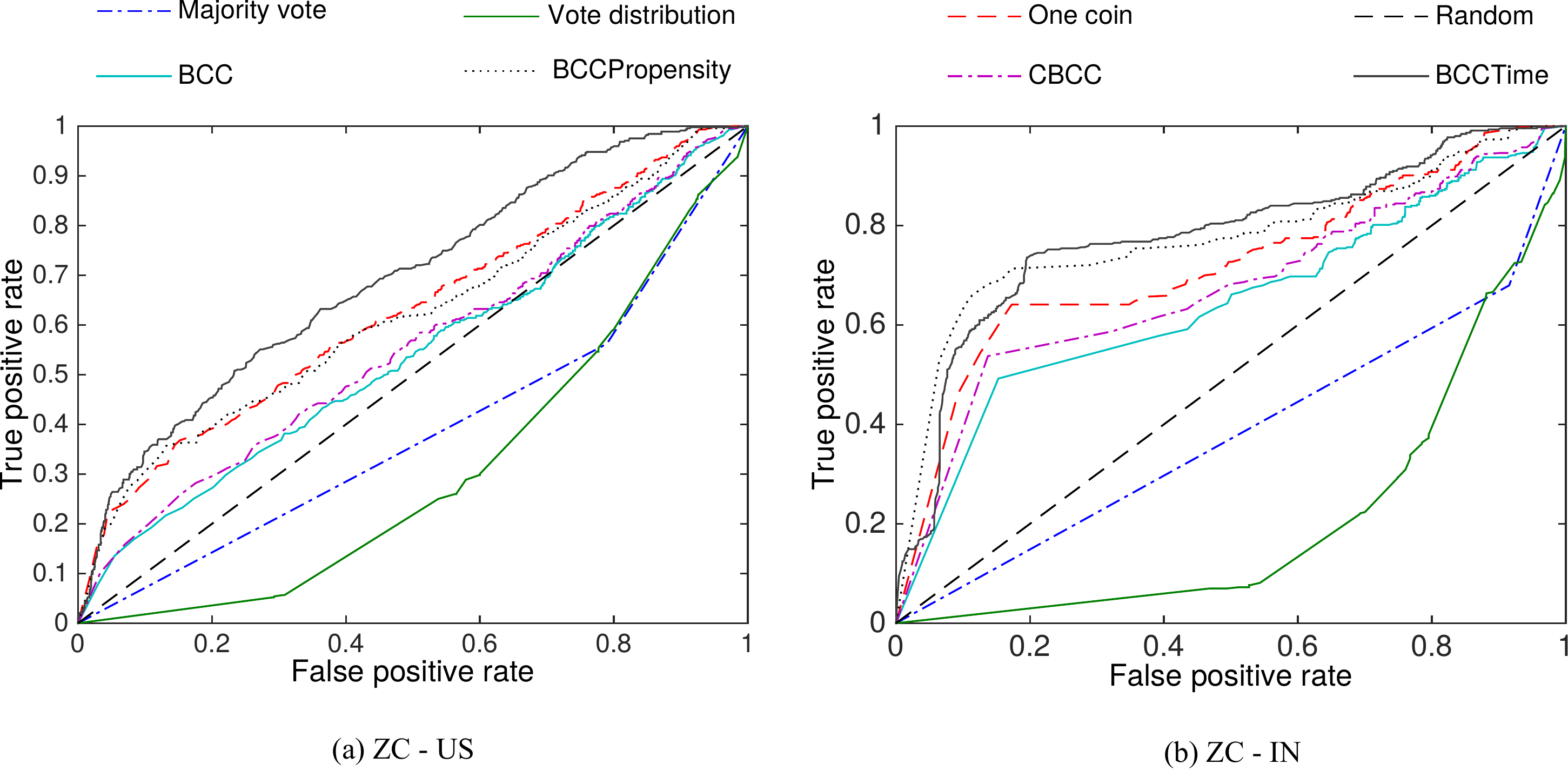}
\caption{The ROC curve of the aggregation methods for ZC-US (a) and ZC-IN (b).}
\label{fig:roc}
\end{figure*}

Figure \ref{fig:roc} shows the ROC curve of the methods for the ZenCrowd (binary) datasets, namely the plot of the false positive rate and the true positive rate obtained for different discriminant thresholds. The graph shows that the true positive rate of BCCTime is generally higher than that of the benchmarks at the same false positive rate.
In detail, Majority vote, and Vote distribution perform worse than Random in these datasets as these methods are clearly penalised by the presence of less reliable workers as they treat all the workers as equally reliable. Interestingly, One coin performs better than BCC and CBCC meaning that the confusion matrix is better approximated by a single (one coin) parameter for these two datasets.
Also, looking at the percentages of the workers' propensities inferred by BCCTime reported in Table \ref{tab:prop}, we found that $93.2\%$ of the workers  in ZC-US, $60\%$ of the workers in ZC-IN  and $97.3\%$ of the workers in WS-AMT have a propensity greater than $0.5$. This means that, in ZC-US and WS-AMT,  only a few workers were identified as suspected spammers while the majority of them were estimated as more reliable with different propensity values. In ZC-IN, the percentage of suspected spammers is higher and this is also reflected in the lower accuracy of the judgments with respect to the gold standard labels.

\begin{figure*}[h!]
\centering
\includegraphics[width=0.68\textheight]{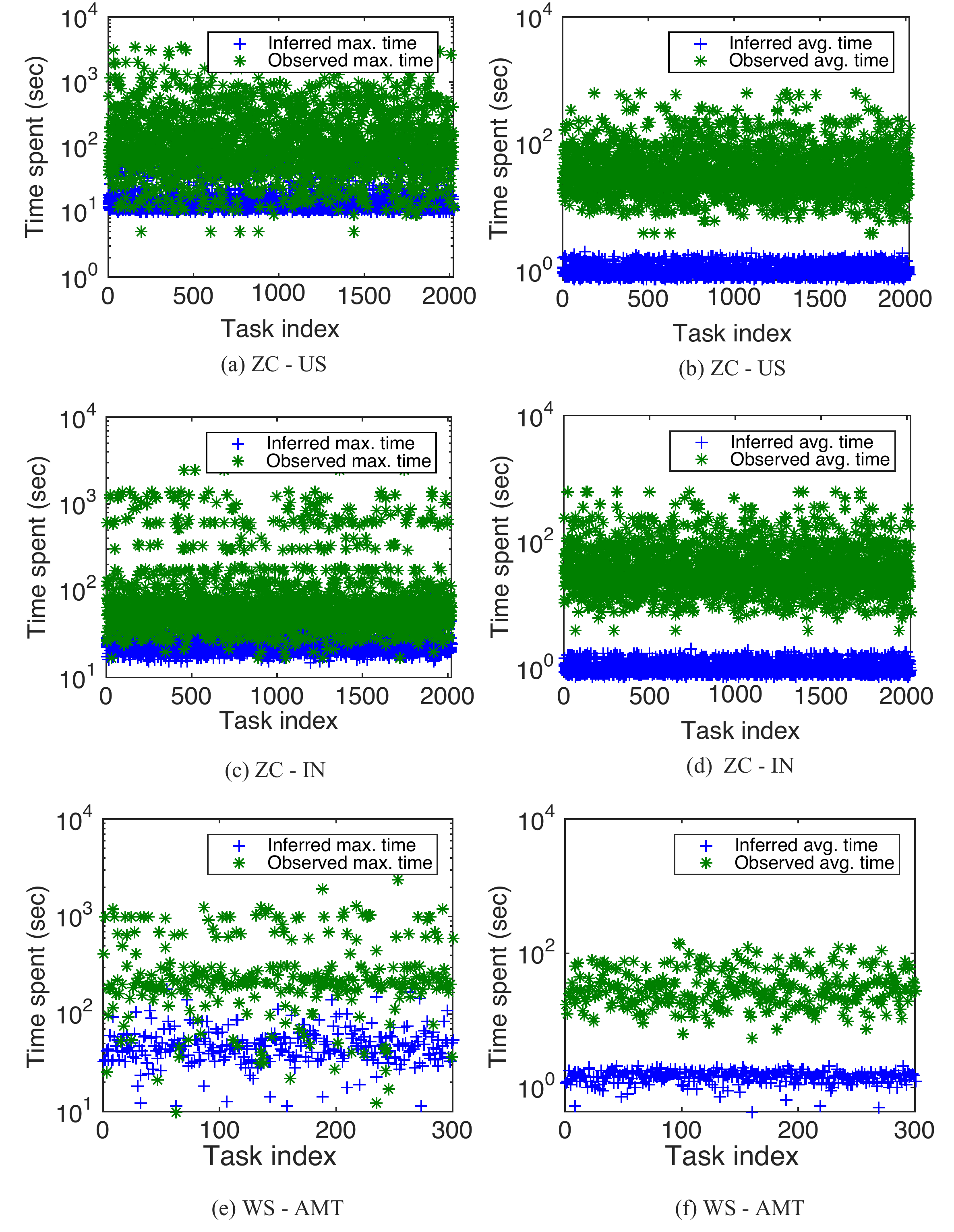}
\caption{The plot of the inferred (+) and observed (*) maximum time spent on the tasks in ZC-US (a), ZC-IN (c) and WS-AMT (e), and the average time spent on the tasks in ZC-US (b), ZC-IN (d) and WS-AMT (f).}
\label{fig:res_time1}
\end{figure*}

Figure \ref{fig:res_time1} shows the mean value of the inferred upper-bound time threshold $\lambda_i$ (blue cross points) and the workers' maximum completion time (green asterisked points) for each task of the three datasets.
Looking at the raw data in the ZenCrowd datasets, the average maximum time spent by the US workers is higher (approx. 1.7 minutes) than that of the Indian workers (approx. 1 minute).
It can also be seen that in both datasets there is a significant portion of outliers that reach up to 50 minutes.
However, as discussed in Section \ref{sec:analysis}, we know that many of these entity linking tasks are fairly simple -- some of them can easily be solved through visual inspection of the candidate URI. This does not imply that a normal worker who completes the task in a single session (i.e., no interrupts) should take such a long time to issue her judgment.
Interestingly, BCCTime efficiently removes these outliers and recovers more realistic estimates of the maximum duration of an entity linking task. In fact, its estimated upper-bound time thresholds lie within a smaller time band, i.e., around 10 seconds in ZC-US and 40 seconds in ZC-IN.
 Similar results are also observed in WS-AMT where the average observed maximum time is significantly higher than the average inferred maximum time, thus suggesting that the BCCTime estimates are also more realistic in this dataset.
 In addition, Figure \ref{fig:res_time1} shows the same plot for the average duration as estimated by BCCTime (i.e, ${(E[\lambda_i] - E[\alpha_i])}/{2} \quad \forall i$) and the average worker's completion time for each task. The graphs show that the BCCTime estimates are similar between the micro-tasks of the three datasets, i.e., between 3 and 5, while the same estimates obtained from the worker's completion time data are much higher: 53 seconds for ZC-US, 45 seconds for ZC-IN and 80 seconds in WS-AMT. Again, this is due to the presence of outliers in the original data that significantly bias the empirical average times towards high values.
Moreover, measuring the variability in the two sets of estimates, the BCCTime estimates have a much smaller standard deviation that is up to $100\%$ lower than that of the empirical averages.
 This means that our estimates are more informative when compared to the normal average times obtained from the raw workers' completion time data.
 
\begin{figure*}[t]
\centering
\includegraphics[width=\textwidth]{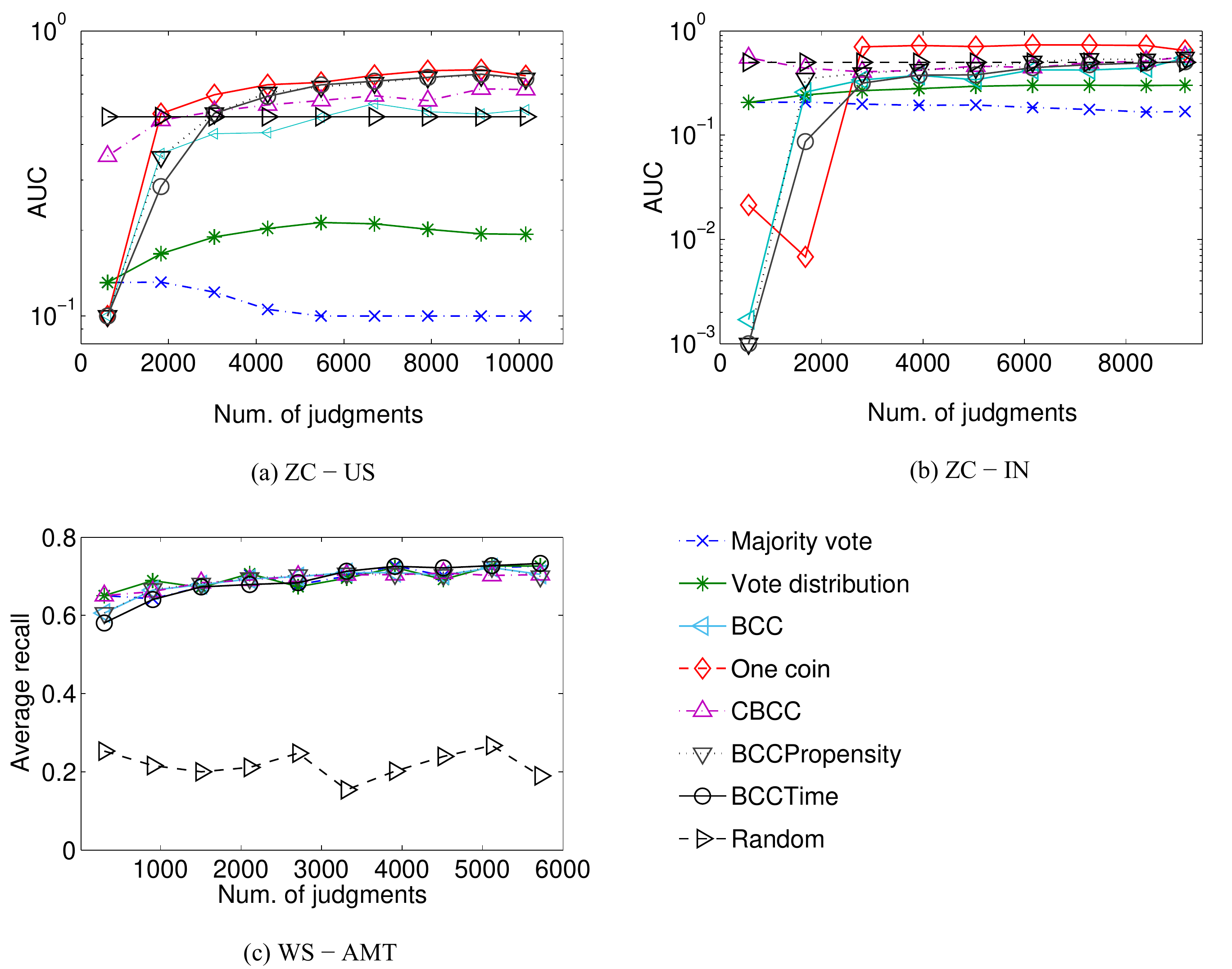}
\caption{The AUC in ZC-US (a) and ZC-IN (b) and the average recall in WS-AMT (c) of the methods trained over increasingly large sub-sets of judgments.}
\label{fig:res_active_learning}
\end{figure*}

 To evaluate the performance of the methods against data sparsity, Figure \ref{fig:res_active_learning} shows the accuracy measured over sub-samples of judgments in each dataset. In more detail,  one coin is more accurate over sparse judgments in ZC-IN and ZC-US, while in WS-AMT there is no clear winner since all the methods except Random have a similar average recall when trained on sparse judgments. This shows that BCCTime in the current form does not necessarily outperform the other methods with sparse data. This can be explained by the fact that the extra latent variables (i.e., workers' propensity and time thresholds) used to improve the quality of the final labels also require a larger set of judgments to be accurately learnt. However, to address this issue, it is possible to draw from community-based models (e.g., CBCC) to design a hierarchical extension for BCCTime over, for example, the workers' confusion matrices and so improve its robustness on sparse data.
Here, for simplicity, BCCTime is presented  based on  simpler instance of Bayesian classifier combination framework (i.e., the BCC model), and  its community-based version is considered as a trivial extension.

\section{Related Work}
\label{sec:related}
Here we review the rest of previous work relating to aggregation models and time analysis in crowdsourcing contexts extending the background of the methods already considered in our experimental evaluation.
In recent years, a large body of literature has focussed on the development of smart data aggregation methods to aid requesters in combining judgments from multiple workers.
In general, existing methods vary by assumptions and complexity in modelling the different aspects of labelling noise.
The interested reader  may refer to the survey by \citeauthor{sheshadri2013square} (2013), as well as to the summary in Table \ref{tab:list} that lists the most popular methods and their comparison with our approach.

In particular, some of these methods are able to handle both binary classification problems, i.e.,  when workers have to vote on objects between two possible classes, and multi-class classification problems, i.e., when workers have to vote on objects between more than two classes.
Among these, many approaches use the one coin model introduced in our benchmarks. In more detail, this model represents the worker's reliability with a single  parameter defined within the range of $[ 0, 1]$ (0 = unreliable worker, 1 = reliable worker) \cite{karger2011iterative,liu2012variational,demartini2012zencrowd,li2014wisdom,nushi2015crowd}.
Specifically, \cite{karger2011iterative} combines this model with a budget--limited task allocation framework and provides strong theoretical guarantees on the asymptotical optimality of the inference of the workers' reliability and the worker-task matching.
 \cite{liu2012variational} uses a more general variational inference model that reduces to \citeauthor{karger2011iterative}'s method, as well as other algorithms under special conditions.
Other methods use a two coin model that represents the bias of a worker towards the positive labelling class (specificity) and towards the negative class (sensitivity) \cite{raykar2010learning,rodrigues2014gaussian,bragg2013crowdsourcing}. Then,  these quantities may be inferred  using logistic regression as in \cite{raykar2010learning} or maximum--a--posteriori approaches as in \cite{bragg2013crowdsourcing}.
Alternatively, \cite{rodrigues2014gaussian} uses the two coin model embedded in a  Gaussian process classification framework to compute the predictive probabilities of the aggregated labels and the workers' reliability using EP.
Along the same lines, other models reason about the difficulty of a task that affects the quality of a judgment to improve the reliability of aggregated labels \cite{whitehill2009whose,bachrach2012grade,kajino2012convex}.
In this area, \cite{whitehill2009whose}  use a logistic regression model to incorporate the task's difficulty, together with the expertise of the worker for labelling images.
In contrast, \cite{bachrach2012grade} use the difference between these two quantities to quantify the advantage that the worker may have in classifying the object within a joint difficulty-ability-response model.
In a similar setting, \cite{kajino2012convex} exploit a convex problem formulation of this model to improve the efficiency of inferring these quantities through a numerical optimisation method.
Additional factors, such as the worker's motivation or propensity for a particular task, are taken into account in more sophisticated models introduced by  \cite{welinder2010multidimensional,yan2010modeling,bilearning}.
More recently, \cite{nushi2015crowd} devised a method that leverage the fact that the error rates of the workers are directly affected by the access path they follow, where the access path represents several contextual features of the task (e.g., task design, information sources and task composition).
However, unlike our work, none of these methods learn the confusion matrix of each worker. As a result, they do not represent reliability considering the accuracy and the potential biases of a worker with a single data structure.

Alternative models that do learn the confusion matrices of the workers have been presented, among others, by \cite{dawid1979maximum,zhou2012learning,kim2012bayesian,venanzi2014community}.
In particular, \cite{dawid1979maximum} introduced the first confusion matrix-based model in which the confusion matrices are inferred using expectation-maximisation in an unsupervised manner.
Then, \cite{zhou2012learning} extended this work to include a task--specific latent matrix representing the confusability of a task as perceived by the workers.
However, neither of these methods consider the uncertainty over the worker's reliability and the other parameters of their models. For example, when only one label is obtained from a worker, these methods may infer that the worker is perfectly  reliable or totally incompetent when, in reality, the worker is neither.
To overcome this limitation, other methods such as BCC  and CBCC capture the uncertainty in the worker's expertise and the true labels using a Bayesian learning framework. These two methods were extensively discussed earlier (see Sections \ref{sec:prel} and \ref{sec:exp}) and are included as benchmarks in our experiments.
Similarly to CBCC,  other methods leverage groups of workers with equivalent reliability to improve the quality of the aggregated labels with limited data \cite{li2014wisdom,bilearning,kajino2012convex,yan2010modeling}.
However, as already noted, all these methods do not use any extra information other than the workers' judgments to learn their probabilistic models. As a result, unlike our approach, they cannot take full advantage of the time information provided by the crowdsourcing platform to improve the quality of their inference results.

\begin{sidewaystable}[ph!]
\caption{Comparison of 21 existing methods for computing aggregated labels from crowdsourced judgments.}
\vspace{0.1cm}
\centering
\begin{tabular}{lccccccc}
\hline
\hline
&binary&multi&worker&worker&task&task&worker\\
&class&class&accuracy&confusion matrix&difficulty&duration&type or propensity\\
\hline
Majority voting&\checkmark&\checkmark&-&\xmark&\xmark&\xmark&\xmark\\
\hline
DS - Dawid \& Skene (1979)&\checkmark&\checkmark&\checkmark&\checkmark&\xmark&\xmark&\xmark\\
\hline
GLAD - Whitehill et al. (2009)&\checkmark&\xmark&\checkmark&\xmark&\checkmark&\xmark&\xmark\\
\hline
RY - Raykar et al. (2010)& \checkmark&\xmark&\checkmark&\xmark&\xmark&\xmark&\xmark\\
\hline
CUBAM - Welinder et al. (2010)&\checkmark&\xmark&\checkmark&\xmark&\xmark&\xmark&\checkmark\\
\hline
YU - Yan et al. (2010)&\checkmark&\checkmark&\checkmark&\xmark&\xmark&\xmark&\xmark\\
\hline
LDA - Wang et al. (2011)&\xmark&\xmark&\xmark&\xmark&\xmark&\checkmark&\xmark\\
\hline
KJ - Kajino et al. (2012)&\checkmark&\xmark&\checkmark&\xmark&\xmark&\xmark&\checkmark\\
\hline
ZenCrowd - Demartini et al. (2012)& \checkmark&\xmark&\checkmark&\xmark&\xmark&\xmark&\xmark\\
\hline
DARE - Bachrach et al. (2012)&\checkmark&\checkmark&\checkmark&\xmark&\checkmark&\xmark&\xmark\\
\hline
MinMaxEntropy - Zhou et al. (2012)&\checkmark&\checkmark&\checkmark&\checkmark&\xmark&\xmark&\xmark\\
\hline
BCC - Kim \& Ghahramani (2012)&\checkmark&\checkmark&\checkmark&\checkmark&\xmark&\xmark&\xmark\\
\hline
MSS - Qi et al. (2013)&\checkmark&\checkmark&\checkmark&\xmark&\xmark&\xmark&\checkmark\\
\hline
MLNB - Bragg et al. (2013)&\checkmark&\checkmark&\checkmark&\xmark&\xmark&\xmark&\xmark\\
\hline
BM - Bi et al. (2014)&\checkmark&\xmark&\checkmark&\xmark&\checkmark&\xmark&\xmark\\
\hline
GP - Rodriguez et al. (2014)&\checkmark&\xmark&\checkmark&\xmark&\xmark&\xmark&\xmark\\
\hline
LU - Liu et al. (2014)&\checkmark&\xmark&\checkmark&\xmark&\xmark&\xmark&\xmark\\
\hline
WM -Li et al. (2014)&\checkmark&\checkmark&\checkmark&\xmark&\xmark&\xmark&\checkmark\\
\hline
CBCC - Venanzi et al. (2014)&\checkmark&\checkmark&\checkmark&\checkmark&\xmark&\xmark&\checkmark\\
\hline
APM - Nushi et al. (2015)&\checkmark&\checkmark&\checkmark&\xmark&\xmark&\xmark&\xmark\\
\hline
BCCTime - \emph{Proposed method}&\checkmark&\checkmark&\checkmark&\checkmark&\checkmark&\checkmark&\checkmark\\
\hline
\hline
\end{tabular}
\label{tab:list}
\end{sidewaystable}

Now we turn to the problem of time analysis in crowd generated content. recently introduced a metric for measuring the effort required to complete a crowdsourced task based on the area under the error-time curve (ETA). As such, this metric supports the idea of considering time as an important factor a crowdsourcing effort.
In this regard, a closely related work on the analysis of the ZenCrowd datasets (see Section \ref{sec:analysis}) was presented by \cite{difallah2012mechanical}. Their work showed that workers who complete their tasks too fast or too slow are typically less accurate than the others. These findings were also confirmed in our work. However, in addition, we extended their analysis by showing the judgment's quality is correlated to the time spent by the workers in different ways for specific task instances. This is the intuition that our method exploits to efficiently combine the workers' completion time features in the data aggregation process.
Furthermore, earlier work introducing a method that predicts the duration of the task based on a number of available features (including the task's price, the creation time and the number of assignments) using a survival analysis model was presented by  \cite{wang2011estimating}. However, their method does not deal with aggregating  labels, nor learning the accuracy of the workers, as we do in our approach.

\section{Conclusions}
\label{sec:conc}
We presented and evaluated BCCTime, a new  time--sensitive aggregation method that simultaneously merges crowd labels and estimates the duration of individual task instances using principled Bayesian inference.
The key innovation of our method is to leverage an extended set of features comprising the workers' completion time and the judgment set. When appropriately correlated together, these features become important indicators of the reliability of a worker that, in turn, allow us to estimate the final labels, the tasks' duration and the workers' reliability more accurately.
Specifically, we introduced a new representation of the accuracy profile of a worker consisting of both the worker's confusion matrix, which accounts for the worker's labelling probabilities in each class, and the worker's propensity to valid labelling, which represents the worker's intention to meaningfully participate in the labelling process.
Furthermore, we used latent variables to represent the duration of each task using pairs of latent thresholds to capture the time interval in which the best judgments for that task are likely to be submitted by honest workers.
 In this way, the model can deal with the differences in the time length of each task instance relating to the different type of correlation between quality of the received judgments and the time spent by the workers. In fact, such task--specific correlations have been observed in our experimental analysis of crowdsourced datasets in which various task instances showed different types of quality--time trends. Thus, the main idea behind BCCTime is to model these trends in the aggregation of crowd judgments to make more reliable inference about all the quantities of interest.
Through an extensive experimental validation on real-world datasets, we showed that BCCTime produces significantly more accurate classifications and its estimates of the tasks' duration are considerably more informative than common heuristics obtained from the raw workers' completion time data.

Against this background, there are several implications of this work concerning various aspects of reliable crowdsourcing systems. Firstly, the process of designing the task can take exploit the unbiased task's duration estimated by BCCTime. As we have shown, this information is a valid proxy to assess the difficulty of a task and therefore supports a number of decision--making problems such as fair pricing for more difficult tasks and defining fair bonuses to honest workers. Secondly, the worker's propensity to valid labelling uncovers an additional dimension of the workers' reliability that enables us to score their attitude towards correctly approaching a given task. This information is useful to select different task designs or more engaging tasks for workers who systematically approach a task incorrectly.
Thirdly, our method uses only features that are readily available in common crowdsourcing systems, which allows for a faster take up of this technology in real applications.

Building on these advances, there are  several aspects of our current model that indicate promising directions for further improvements. For example, we can consider that time--dependencies in the  accuracy profile of a worker capture the fact that workers typically improve their skills over time by performing a sequence of tasks. By so doing, it is possible to take advantage of these temporal dynamics to potentially improve the quality of the final labels. In addition, some crowdsourcing settings  involve continuous-valued judgments that are currently not supported by our method. To deal with these cases, a number of non--trivial extensions to our generative model and, in turn, a new treatment of its probabilistic inference are required.

\newpage

\bibliography{cs_refs}
\bibliographystyle{theapa}

\end{document}